%% file: main.tex
\PassOptionsToPackage{table}{xcolor}

\documentclass{article}
\usepackage{conference,times}

\input{math_commands.tex}

\usepackage{hyperref}
\usepackage{url}

\usepackage{graphicx}
\usepackage{booktabs}
\usepackage{longtable}
\usepackage{array}
\usepackage[accsupp]{axessibility}
\usepackage{amsmath}
\usepackage{wrapfig}
\usepackage{multirow}
\usepackage{float}
\usepackage{placeins}
\usepackage{enumitem}
\newcommand{\keywords}[1]{} 
\usepackage{graphicx}
\usepackage{subcaption}
\newcommand{\framework}{DexLink}
\definecolor{AccentBrown}{HTML}{A65A2A}
\definecolor{TitleBrown}{HTML}{76513D}
\hypersetup{
    colorlinks=true,
    linkcolor=AccentBrown,  
    citecolor=AccentBrown,  
    urlcolor=AccentBrown   
}
\input{notations}

\title{\framework: Synthesizing Dexterous Behaviors from Chained Hand-object Poses}
\author{
Meenal Parakh$^\dagger$,
Deep Patel$^\S$,
Beining Han$^\dagger$,
Jia Deng$^\dagger$ \\
$^\dagger$Princeton University \\
$^\S$NEC Laboratories America
}

\conferencefinalcopy

\usepackage{xcolor}
\colorlet{revisioncolor}{black}
\newcommand{\revised}[1]{{\color{revisioncolor}#1}}

\begin{document}

\maketitle

\begin{abstract}

Synthesizing diverse dexterous manipulation policies is challenging: existing simulation-based approaches rely on human demonstrations~\citep{jiang2025dexmimicgen, mu2026deximit} which are difficult to scale, or use generic reward functions~\citep{kedia2026simtoolreal}, which provide limited control over the hand-object interactions. We propose \framework, which addresses these limitations by synthesizing diverse dexterous behaviors from sparse, reusable hand-object interaction templates. Given key task waypoints, \framework~retargets static interaction templates to target objects using an optimization that preserves
spatial hand-object relationships. It then links the retargeted poses into an approximate reference trajectory and uses residual reinforcement learning to refine this reference
into a closed-loop policy in simulation. Experiments spanning single-hand, bimanual, articulated, and dynamic tasks demonstrate that \framework~provides fine-grained control over
task timing, grasp types, and high-level execution. The modular interaction templates transfer reliably across object scales, instances, and categories; without object-specific
human demonstrations, the framework also matches the performance of dense reference-tracking methods. Finally, zero-shot deployment of policies trained purely in simulation shows promise for the simulation-based framework. 
\href{https://dex-link.github.io/}{\texttt{https://dex-link.github.io/}}
\end{abstract}

\section{Introduction}
\vspace{-5pt}

Learning from data is a key paradigm in robotic manipulation and relies on the ability to collect large, diverse behaviors \citep{droid}. For dexterous manipulation, this requires coverage over diverse tasks, objects, and interaction behaviors. Obtaining such coverage is challenging: dexterous hands involve complex, contact-rich dynamics that make teleoperation cumbersome and expensive~\citep{handa2020dexpilot, tao2025dexwild}.

Simulation provides a scalable platform for synthesizing dexterous behaviors, with potential for real-world transfer \citep{jiang2025dexmimicgen, chen2023visual}. Yet, specifying which tasks and behaviors to generate, and how, remains an open challenge. Existing work often relies on human demonstrations \citep{jiang2025dexmimicgen, li2025maniptrans} to generate robot data, creating a scaling bottleneck, or limits generation to only specific skills such as grasping~\citep{ye2025dex1b, chen2025dexonomy}. We aim to address these challenges, enabling simulation-based synthesis of general dexterous behaviors.

\begin{figure}[h!]
    \centering
    \includegraphics[width=1\linewidth]{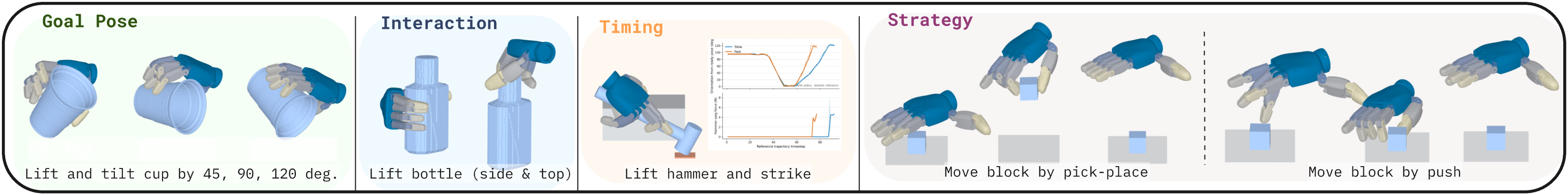}
    \caption{Controllable dexterous behavior axes: goal pose, hand–object interaction, task timing, and high-level strategy.}
    \label{fig:control_axes}
    \vspace{-10pt}
\end{figure}

Towards this goal, we present \framework, a simulation-based framework that enables generation of dexterous behaviors across a diverse set of tasks and objects, spanning single-hand, bimanual, long-horizon, articulated, and dynamic tasks. Further, \framework~allows control over behavior synthesis: object goal poses, hand-object interactions, task timing, and high-level execution strategies (Figure~\ref{fig:control_axes}). 

Importantly, \framework~provides these capabilities without requiring costly human demonstrations or cumbersome per-task reward engineering, which existing works often depend on. Existing generation methods vary in how they input task-related information, from indirect optimization objectives to full task demonstrations. Indirect objectives such as task-specific reward design~\citep{Chen2021ASF, kedia2026simtoolreal} or hand priors~\citep{mandikal2022dexvip, de2026scaffolding} in RL indirectly guide policies toward specific behaviors: while such guidance is easy to obtain, it provides only weak control over learned behaviors and can produce unnatural robot motions or interactions that fail to satisfy downstream task requirements. In contrast, methods such as \citet{jiang2025dexmimicgen, li2025maniptrans, mu2026deximit} provide task supervision using dense demonstrations obtained from teleoperation, motion-capture data, or vision-based annotations, respectively: the dense supervision provides more control over learning but is costly to collect and, when extracted from video, depends on the accuracy of vision modules. Further, the generation is typically tied to the specific task-object pair represented in the demonstration, limiting scalability.

\begin{figure}[h!]
    \centering
    \vspace{-5pt}
    \includegraphics[width=0.8\linewidth]{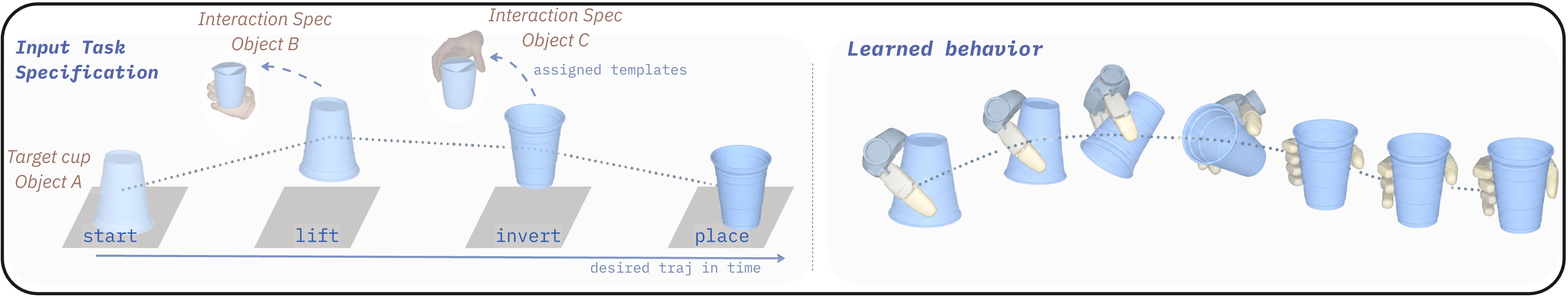}
    \caption{(a) An example task representation provided to the framework. (b) \framework~synthesizes a grounded policy that follows the specification.}
    \label{fig:illustrative-figure}
    \vspace{-10pt}
\end{figure}

\framework~instead achieves the \textit{diversity} and \textit{control} over behavior generation by assuming a sparse, compositional task representation which
combines a sequence of object-state waypoints with hand-object interaction templates assigned to selected waypoints (shown in Figure~\ref{fig:illustrative-figure}, left). The object waypoints describe the desired task progression, while the interaction templates specify the desired spatial relationship between the hand and object. Importantly, these templates only specify local interactions and can be reused across tasks and objects and chained with other templates along the waypoint sequence. Selecting different waypoints and interaction templates allows the target behavior to be varied, or controlled, as desired. These templates can be obtained from motion-capture frames or static hand-object reconstruction methods~\citep{yang2022oakinklargescaleknowledgerepository, fan2023holdcategoryagnostic3dreconstruction}.

Obtaining a physically grounded trajectory from this compositional task representation presents a challenge: we need to preserve the desired interactions, goals, and phase timings, while remaining dynamically feasible.
\framework~addresses this grounding problem by decomposing it into two steps: constructing a reference trajectory that preserves the task intent, and refining the reference for dynamic feasibility.  
Specifically, an object-aware retargeting optimization adapts each interaction template to the target-object geometry, producing approximate robot anchors. These anchors are linked along the waypoints to form an intermediate reference, which is refined with residual RL in simulation. The intermediate reference narrows RL exploration around the intended behavior, while RL corrects its dynamics, contact, and timing errors to provide a physically grounded trajectory (Figure \ref{fig:illustrative-figure} right).

Our experiments, covering a diverse set of tasks, evaluate the framework's flexibility: interaction templates can be reused across scale, within-category and across-category variations, and generation can be carefully controlled by varying the task specification. Ablations establish the importance of each component, and we show that the framework supports sparse interaction anchors, while performing on par with other approaches for dense object tracking. Finally, zero-shot deployment of policies trained purely in simulation demonstrates the potential for zero-shot real-world transfer.

In summary, our contributions are: (i) a simulation-based framework for synthesizing diverse dexterous behaviors across tasks and objects; 
(ii) an approach that converts composable sequences of object-state waypoints and hand-object interaction templates into physically grounded policies;
and (iii) systematic evaluations in simulation demonstrating coverage over variations in object geometries and tasks, and zero-shot real-world deployment of simulation policies.

\section{Related Works}
\vspace{-5pt}

\textbf{Generating Dexterous Demonstrations.} 
Simulation-based approaches specify dexterous behaviors in several ways.
One category of methods specifies desired behaviors via reward functions in RL \citep{Chen2021ASF, mandikal2022dexvip}. While the same reward function can enable learning across diverse objects, generalization to diverse tasks typically requires task-specific reward engineering. Works such as \citet{kedia2026simtoolreal, de2026scaffolding} can scale to different objects and tasks but cannot control lower-level finger-object interaction patterns, which may matter for downstream tasks. Alternately, dense interaction guidance can be obtained through teleoperation \citep{jiang2025dexmimicgen}, annotated human videos \citep{hsieh2025dexman, mu2026deximit}, or MoCap data \citep{mandi2025dexmachina, li2025maniptrans}, avoiding task-specific reward engineering. However, such data is costly to collect and leaves limited room for data augmentation. Generalization to new tasks and objects can be achieved by pretraining on large datasets~\citep{zhao2025resmimicgeneralmotiontracking}, augmenting demonstrations~\citep{chen2024objectcentricdexterousmanipulationhuman}, or iteratively expanding object diversity~\citep{zhu2026dexflywheel}. However, using learned priors for generation itself remains challenging: the generation may not adhere to the conditioning and may generate only dominant modes \cite{dhariwal2021diffusionmodelsbeatgans}, limited by the scale of pretraining. \framework~instead uses sparse interaction templates reused across tasks and objects, while still providing control over interactions at key stages.

\textbf{Generating Dexterous Grasps.} Dexterous grasping is a fundamental component of dexterous behaviors, as many tasks can be decomposed into individual grasp stages \citep{mu2026deximit}. Methods such as \cite{chen2025dexonomy, contact_map_transfer, zhong2026grasp2grasp} enable controlled generation of different grasp types from interaction templates. However, these works cannot be easily extended to dynamic tasks, which involve complex interactions that force-closure constraints \citep{chen2025dexonomy}, or learned generative models \citep{contact_map_transfer} cannot output precisely. Although such methods could be incorporated into \framework~for reference generation, we show that an analytical objective is sufficient to transfer interaction templates into a coarse reference that downstream RL can refine.

\textbf{Human-Robot retargeting.} 
Existing retargeting methods~\citep{qin2023anyteleop, yin2025geometricretargeting, qin2022dexmv} map human hand poses to robot hand configurations by preserving fingertip positions or vectors between hand keypoints, such as Task Space Vectors. However, these objectives do not account for the target object's geometry. 
\framework~instead uses object-aware retargeting to optimize robot hand configurations that preserve the source interaction while conforming to the target object.

\section{\framework}

\begin{figure}[t]
\centering
\includegraphics[width=0.9\textwidth]{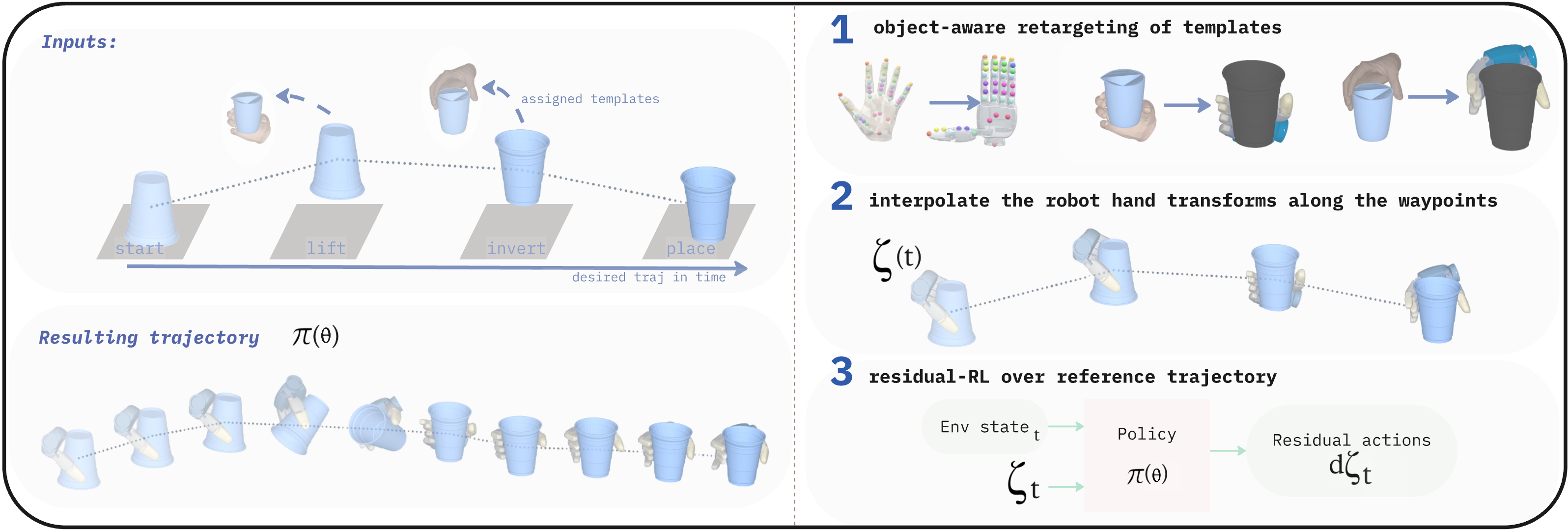}
\vspace{-5pt}
\caption{\framework~overview: Given hand-object pose waypoints, object-aware retargeting computes interaction anchors, which are chained to construct an approximate reference trajectory. Residual RL refines the reference into a closed-loop policy.}
\label{fig:method_pipeline}
\vspace{-5pt}
\end{figure}

Figure~\ref{fig:method_pipeline} provides an overview of the framework. The input is a sequence of object-state waypoints, with hand-object interaction templates assigned to selected waypoints where the desired interaction is important for task success. The output is a closed-loop dexterous policy learned in simulation. The framework consists of two steps: (a) retargeting and composing the selected templates along the task waypoints to construct an approximate reference trajectory, and (b) refining this reference with residual RL.

\subsection{Task Specification} 
The object-state task specification \(\TaskSpec = \{\SceneState_0, \SceneState_1, \ldots, \SceneState_K\}\) is a sequence of desired scene states. Each \(\SceneState_k\) is associated with a time \(t_k\) and specifies the desired poses of relevant objects and, when applicable, their articulation values. These waypoints encode the final goal and any intermediate states required to complete the task.

At selected waypoints \(k\), the task specification includes a human hand-object interaction template \(\HOTemplate_k\) assigned to a target scene object with mesh
\(\ObjMesh^{\mathrm{tgt}}\):
\[
  \HOTemplate_k =
  \left(
  \ObjMesh^{\mathrm{src}}_k,
  P^{\mathrm{h}}_k,
  \SrcToTgtRotK
  \right).
\]
Here, \(\ObjMesh^{\mathrm{src}}_k\) is the source object mesh, \(P^{\mathrm{h}}_k\) contains the 3D human-hand keypoints expressed in the source-object frame, and \(\SrcToTgtRotK \in \SO(3)\) maps the source-object frame to the target-object frame. The source mesh and hand keypoints can be obtained from motion capture or hand-object reconstruction methods. We assume that the chosen templates represent a geometrically meaningful interaction that can be transferred to the target object or a relevant object part, similar to grasp
taxonomies~\citep{chen2025dexonomy}. 

\begin{wrapfigure}{r}{0.35\textwidth}
    \centering
    \vspace{-5pt}
    \includegraphics[width=0.95\linewidth]{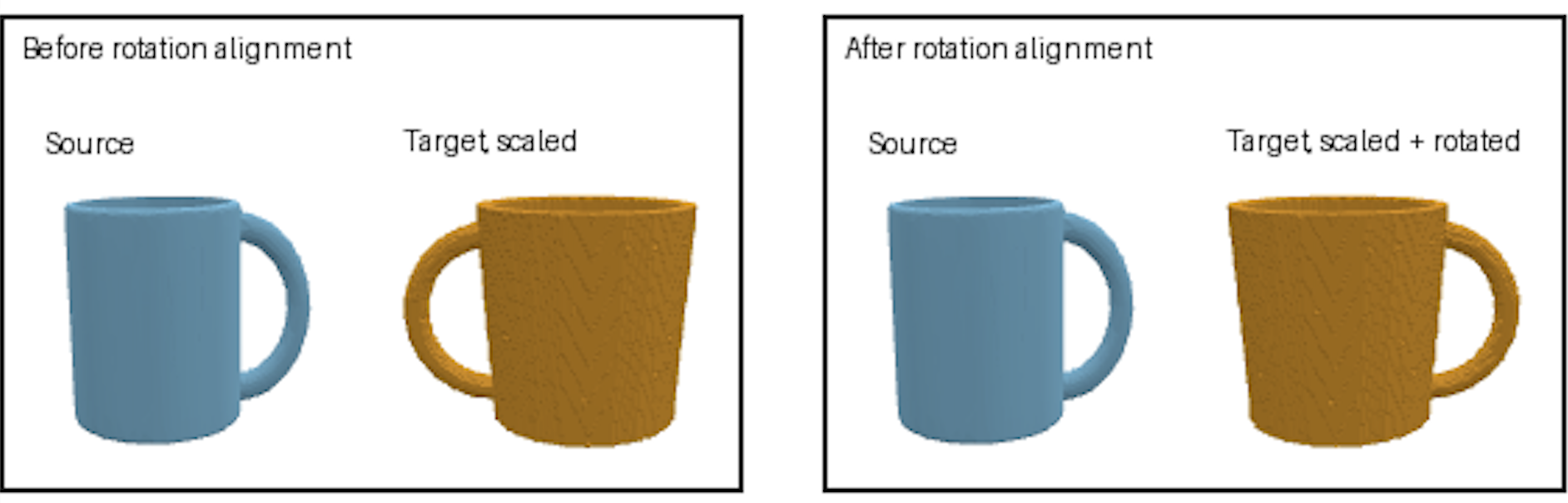}
    \caption{Example of alignment required for task-relevant source and target object parts before retargeting.}
    \vspace{-10pt}
    \label{fig:rot_align}
\end{wrapfigure}

Figure~\ref{fig:rot_align} shows the rotational alignment. Applying \(\SrcToTgtRotK\) expresses the source mesh and hand keypoints in the target-object axes:
\[
\bar{\ObjMesh}_k^{\mathrm{src}}
  = \SrcToTgtRotK \ObjMesh_k^{\mathrm{src}},
\qquad
\bar P_k^{\mathrm h}
  = \SrcToTgtRotK P_k^{\mathrm h}.
\]

\subsection{Reference Trajectory Construction}
\label{sec:ref_traj}

The reference trajectory need not be dynamically feasible and may contain timing and interaction errors, provided it biases residual RL toward the intended behavior~\citep{de2026scaffolding}. We construct this reference by retargeting each interaction template \(\HOTemplate_k\) to the target mesh \(\ObjMesh^{\mathrm{tgt}}\) and interpolating between the resulting robot configurations along the task waypoints (Figure~\ref{fig:method_pipeline}, Steps 1-2).

\paragraph{Object-aware retargeting.}

For each template \(\HOTemplate_k\), retargeting solves for a robot wrist pose \(\wrist_k^\star \in \SE(3)\), expressed in the target-object frame, and a joint configuration \(q_k^\star\) that preserves the source hand-object interaction on the target object. Directly matching hand keypoints is unsuitable when the hand embodiment and object geometry differ. Instead, we preserve relations among hand keypoints  using \(\loss_{\mathrm{cfg}}\) and the spatial relationship between the hand and
object using \(\loss_{\mathrm{rel}}\):
  
\[
(\wrist_k^\star,q_k^\star)
=
\arg\min_{\wrist_k\in\SE(3),\,q_k}
\left[
\loss_{\mathrm{cfg}}
  \left(\wrist_k\!\cdot\!\FK(q_k),\bar P_k^{\mathrm h}\right)
+
\lambda\,
\loss_{\mathrm{rel}}
  \left(
      \wrist_k\!\cdot\!\FK(q_k),
      \ObjMesh_k^{\mathrm{tgt}};
      \bar P_k^{\mathrm h},
      \bar{\ObjMesh}_k^{\mathrm{src}}
  \right)
\right].
\]
Here, \(\FK(q_k)\) denotes the robot hand keypoints returned by forward
kinematics in the wrist frame, while
\(\wrist_k\!\cdot\!\FK(q_k)\) denotes the same keypoints expressed in the
target-object frame.

The configuration term preserves geometric relations between hand keypoints,
including finger directions~\citep{qin2022dexmv} and palm orientation. Let
\(\mathcal C\) denote the set of relations, and let \(\phi_c(P)\) extract
relation \(c\) from hand keypoints \(P\). We define
\[
\loss_{\mathrm{cfg}}(P^{\mathrm r},P^{\mathrm h})
=
\frac{1}{|\mathcal C|}
\sum_{c\in\mathcal C}
d_c\!\left(
  \phi_c(P^{\mathrm r}),
  \phi_c(P^{\mathrm h})
\right),
\]
where \(d_c\) is cosine distance for normalized directions and squared
\(\ell_2\) distance for unnormalized vectors.

For the hand-object relational term, we compute smoothed BPS-based descriptors~\citep{Prokudin2019EfficientLO} for the human--source and
robot--target pairs (Figure~\ref{fig:method_pipeline}, Step 1). For each hand keypoint, the descriptor captures its displacement to the object surface \(D\), the corresponding surface normal \(N\), and a normalized object-coordinate feature \(F\). $r$ and $h$ flags correspond to descriptors for the robot and human hand, respectively.
  
\[
\loss_{\mathrm{rel}}
=
\loss_{\mathrm{dist}}(D^{\mathrm r},D^{\mathrm h})
+
\loss_{\mathrm{dir}}(D^{\mathrm r},D^{\mathrm h})
+
\loss_{\mathrm{normal}}(N^{\mathrm r},N^{\mathrm h};D^{\mathrm h})
+
\loss_{\mathrm{feat}}(F^{\mathrm r},F^{\mathrm h}).
\]

The four terms compare hand-object distance, displacement direction, local surface orientation, and normalized object coordinates, respectively. Further details are in the appendix.

Because the descriptor captures spatial hand-object relationships, we can retarget interactions even when the hand and object are not in contact, as in handover tasks. The resulting solution \(q_k^\star\) is made penetration-free for simulation.

\paragraph{Chaining waypoints into a reference.}
Object-aware retargeting produces
\((\wrist_k^\star,q_k^\star)\) at each waypoint with an interaction template.
Consider two successive templated waypoints \(k\) and \(\ell\), associated with
the same object or articulated part, and let
\(\alpha_t=(t-t_k)/(t_\ell-t_k)\) for \(t\in[t_k,t_\ell]\). We construct the
hand reference as
\[
\begin{aligned}
\wrist_t^{\mathrm{ref}}
&=
T_t^{\mathrm{obj}}\!\cdot
\Interp_{\SE(3)}
\left(\wrist_k^\star,\wrist_\ell^\star;\alpha_t\right),\\
q_t^{\mathrm{ref}}
&=
\Interp
\left(q_k^\star,q_\ell^\star;\alpha_t\right).
\end{aligned}
\]

Here \(T_t^{\mathrm{obj}}\) is the interpolated world pose of the corresponding object or part. We interpolate object translations and articulation values linearly, while using spherical interpolation for rotations.
Interpolating the wrist transform in the object frame approximately preserves the specified hand-object relationship as the object moves. We apply this process independently for each hand (in case of bimanual tasks). Finally, the constructed trajectory is processed to remove penetrations. 
The result is a time-indexed reference containing the desired scene state, wrist pose, and finger configuration (Figure \ref{fig:method_pipeline}, Step 2):
\[
    \RefTraj
    =
    \left\{
      \left(\SceneState_t^{\mathrm{ref}},
            {\wrist}_t^{\mathrm{ref}},q_t^{\mathrm{ref}}\right)
    \right\}_{t=0}^{T}.
\]

\subsection{Residual RL refinement} 
The constructed reference provides a nominal motion but may contain errors in
timing, contact, or interaction. We therefore train a closed-loop residual
policy \(\pi_\theta(\cdot;\RefTraj)\) with PPO \citep{schulman2017proximalpolicyoptimizationalgorithms} in simulation, conditioned on \(\RefTraj\), to refine the reference into physically grounded behavior.

At control step \(t\), the policy observes robot proprioception, contact signals,
wrist-relative object states, the current reference state
\(\RefTraj_t\), and the normalized task progress. It predicts bounded corrections
to the reference wrist pose and finger configuration:
\[
  (\Delta \wrist_t,\Delta q_t)
  \sim \pi_\theta(x_t),
  \qquad
  \wrist_t^{\mathrm{cmd}}
  =
  \wrist_t^{\mathrm{ref}}\oplus\Delta\wrist_t,
  \qquad
  q_t^{\mathrm{cmd}}
  =
  q_t^{\mathrm{ref}}+\Delta q_t,
\]

where \(x_t\) denotes the policy observation, and \(\oplus\) represents an addition in $\SE(3)$. The commanded wrist and finger
targets are executed with a low-level PD controller over the joints. 
The reward function for RL consists of: binary task signals, reference tracking, preservation of specified interactions, and motion smoothness:

\[
R_t
=
\sum_{k\in\mathcal K}\bar w_k r_k(t)
+
w_{\mathrm{succ}}\mathbf 1_{\mathrm{success}}(t),
\qquad
\mathcal K
=
\{\mathrm{obj},\mathrm{int},\mathrm{ref},\mathrm{smooth}\}.
\]
where \(\bar w_k\) are normalized dense-term weights. The key dense terms are

{\scriptsize
\[
\begin{alignedat}{2}
r_{\mathrm{obj}}(t)
&=
\frac{1}{|\mathcal O_t|}
\sum_{o\in\mathcal O_t}
\exp\!\left[-\left(
s_p e^p_{o,t}
+s_R e^R_{o,t}
+s_a e^a_{o,t}
\right)\right],
\qquad&
r_{\mathrm{ref}}(t)
&=
\frac{1}{|\mathcal H|}
\sum_{h\in\mathcal H}
\exp\!\left[-s_{\mathrm{ref}}\left(
e^{w,p}_{h,t}
+e^{w,R}_{h,t}
+e^f_{h,t}
\right)\right],
\\[2pt]
r_{\mathrm{int}}(t)
&=
\frac{1}{|\mathcal H|}
\sum_{h\in\mathcal H}
\exp\!\left[-s_{\mathrm{int}}\left(
\lambda_{\mathrm{cfg}}\mathcal L_{\mathrm{cfg}}^h(t)
+
\lambda_{\mathrm{rel}}\mathcal L_{\mathrm{rel}}^h(t)
\right)\right].
&&
\end{alignedat}
\]
}

\(\mathcal O_t\) is the set of tracked objects and
\(\mathcal H\) is the set of robot hands.
The errors \(e^p,e^R,e^a\) measure object position, orientation, and
articulation relative to the reference scene trajectory, while
\(e^{w,p},e^{w,R},e^f\) measure wrist and finger deviation from the robot
reference. The interaction terms
\(\mathcal L_{\mathrm{cfg}}\) and \(\mathcal L_{\mathrm{rel}}\) reuse the
configuration and hand-object objectives introduced in Section \ref{sec:ref_traj}.
\(r_{\mathrm{smooth}}\) is an exponential penalty on the change in joint values in consecutive steps.
Consequently, \(r_{\mathrm{obj}}\) encourages task progress,
\(r_{\mathrm{int}}\) preserve the specified
interaction, \(r_{\mathrm{ref}}\) provides a nominal motion bias, and
\(r_{\mathrm{smooth}}\) discourages abrupt joint motion.

To parallelize learning across long task horizons, each environment samples an initial reference index and initializes the robot and objects from the corresponding collision-refined reference state. This reset distribution allows different portions of the task to be learned concurrently. 

\section{Experiments}
\vspace{-5pt}
We evaluate whether \framework~provides a useful interface for generating diverse dexterous behaviors. Our experiments address the following questions:
\textbf{(A)} Does object-aware retargeting preserve interaction diversity, staying close to the specified human-object interactions?
\textbf{(B)} Can the learned dexterous behaviors be reliably controlled with changes in the task specification, and can interaction templates be reused across variations in object geometry?
\textbf{(C)} How important is the intermediate reference trajectory, and how do other design choices affect policy learning?
\textbf{(D)} How does waypoint density affect policy learning? Does the framework's performance remain on par with dense reference-tracking methods?
\textbf{(E)} Do the learned policies show potential for zero-shot sim-to-real transfer?

\textbf{Setup.} We use the XHand \citep{xhandrobotarm} embodiment in all experiments. Interaction templates are drawn from OakInk \citep{yang2022oakinklargescaleknowledgerepository}, which provides diverse static and functional hand-object interactions, and ARCTIC \citep{fan2023arcticdatasetdexterousbimanual}, which provides temporally dense, bimanual interactions. We use the SNOPT solver in Drake \citep{drake} for object-aware retargeting, and RL policies are trained in the Genesis simulator \citep{Genesis} with 1,024 parallel environments. Unless stated otherwise, each policy is trained for 50k policy rollout steps, corresponding to 51.2M env transitions, using the same policy architecture and optimizer settings. The experiments accumulate results across 3 seeds, selecting checkpoints based on the best logged train-time task metrics. The checkpoints are evaluated with greedy action selection for 50 episodes. RL runs for dense reference tracking experiments use a different training budget and checkpoint selection.

\textbf{Metrics.} Because the tasks differ in their objectives, we use task-specific \textit{success criteria} as the primary metric for each task. We report additional metrics for measuring the quality of rollouts: (i) \textit{Dynamic Time Warping} (DTW, $\downarrow$) between the policy and reference trajectories; and (ii) \textit{ADD-AUC} for dense object tracking, which measures deviation from the reference motion using mesh-vertex tracking accuracy, computed as the area under the success curve across distance thresholds \cite{mandi2025dexmachina}. 

\vspace{-2pt}

\subsection{Results}
\vspace{-2pt}
\textbf{Object-aware retargeting.} The retargeted configurations provide interaction anchors for downstream policy learning. We therefore evaluate whether they are physically plausible, remain faithful to the source interactions, and preserve interaction diversity.

We evaluate 287 OakInk hand-object interactions spanning 33 object categories and retarget each interaction to XHand while keeping the object geometry fixed. We compare object-aware retargeting with two baselines: (i) fingertip-only retargeting, which uses IK with explicit fingertip-position targets; and (ii) fingertip retargeting augmented with a hand-configuration matching cost~\citep{qin2022dexmv}. Both baselines assume that the object geometry remains unchanged and therefore serve as oracle references in this setting.
  
Object-aware retargeting achieves force-closure feasibility in $\sim60\%$ of cases at a $1\text{cm}$ contact threshold (and $80\%$ at a $2\text{cm}$ threshold), showing that retargeted interactions are physically plausible but lack accuracy. On the feasible set, we compute the recall of retargeted interactions for each object mesh, achieving $\sim0.8$ recall over random chance ($0.4$), indicating that the retargeted behaviors remain close to the original specified interactions. Finally, we visualize the spread of feasible retargetings over the human-fitted PCA space (using joint angle vector). We see that retargetings remain well-distributed, and that the optimization does not collapse the diversity of the source interactions.

\begin{figure}[h!]
    \centering
    \includegraphics[width=1\linewidth]{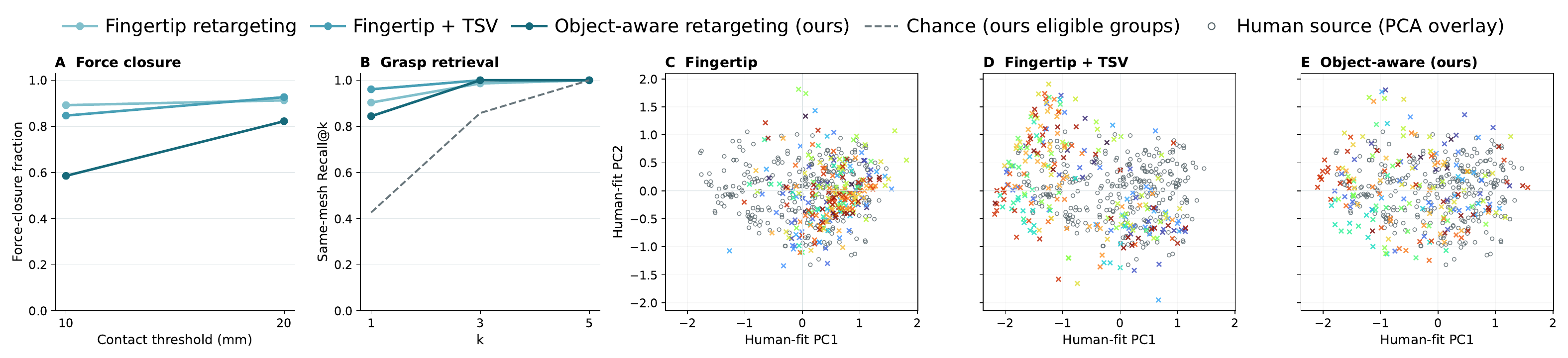}
    \caption{
Quality and diversity of retargeted interactions. (A) Fraction of retargetings satisfying force closure at contact thresholds 1 and 2 cm. (B) Retrieval of retargeted interactions using hand contact map similarity compared to the source template, reported with Recall\@k; dashed line is random retrieval success. (C-E) Feasible robot joint configurations projected onto the first two principal components fitted to the 287 human-hand configurations. Gray circles denote human source templates and colored crosses denote the robot retargetings.}
    \label{fig:retgt_plots}
\end{figure}

\textbf{Robustness to object-geometry changes.} The previous experiment fixed the object geometry across retargeting. We now show that interactions transfer reliably across object-geometry variations, specifically targeting the \texttt{lift} task: grasping and raising the object $15cm$ off the table. The task specification contains two
  waypoints-object on the table and object in the air-with the same interaction template assigned to both. For scale robustness, we transfer five source templates spanning bottle and mug
  interactions to different target instances, each evaluated at four scales. For instance- and category-level transfer, we compare templates transferred from another object or category with templates obtained on the target object. Tables~\ref{tab:scale-reuse} and~\ref{tab:across-category} report mean lift success.

\input{tables/template_transfer_tables}

Overall, we observe perfect lift success on 13/20 scale conditions, concentrated around the nominal scale. Performance
is less reliable at extreme scales and varies across RL seeds; for example, the bottle-side grasp becomes unstable at a target-object scale of \(1.25\). Transferred templates also achieve consistent success across the instance and category-level evaluations (Table~\ref{tab:across-category}), performing comparably to target-specific templates. These results show that a source interaction can guide lifting across changes in object instance, scale, and
category without requiring a separately generated grasp for each target.

\begin{figure}[h!]
    \centering
    \includegraphics[width=0.9\linewidth]{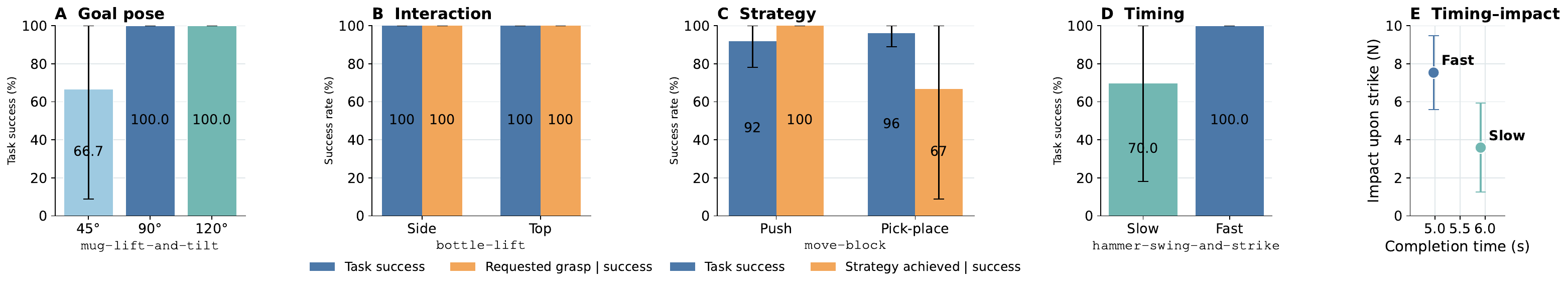}
    \caption{Controllability through task specification. (A) Success for different target mug orientations.   (B-C) Task success and strategy-achieved rate (on success traj.). (D-E) Hammer strike success under slow and fast timing, and the effect of timing on impact upon strike.}
    \label{fig:control_bars}
\end{figure}

\textbf{Task-specification controlling policy behavior.}
We evaluate control over learned behavior along four axes (Figure~\ref{fig:control_axes}). \textit{Goal pose} for \texttt{lift-and-tilt} changes the mug's final tilt angle; \textit{Interaction} changes the grasp template for bottle lift (covered previously); \textit{Timing} for \texttt{hammer-lift-and-strike} alters the timestep of hammer swing waypoints, mainly changing the swing speed; and \textit{Strategy} for \texttt{move-block} alters the intermediate waypoint pose for block, resulting in push and pick-place strategies. Each task variant uses a single interaction template. Figure~\ref{fig:control_bars} reports task success based on the final object state.

We observe that an interpretable task specification interface can generate diverse variations, with control over goal pose, task strategy, and timing: all task variations achieve $> 66\%$. For the \texttt{move-block}, the desired strategy is achieved in $2/3$ seeds, and timing control in the hammer task visibly alters the strength of impact, from $\sim4N$ for slow motion to $\sim7.5N$ for fast motion. Some task variants exhibit high variance across seeds, with two seeds succeeding and one failing completely.

\textbf{Ablations across representative tasks}
 We evaluate the framework's design choices on four representative tasks:  a grasp transition (\texttt{mug-grasp-change}), articulation (\texttt{open-notebook}), sequential manipulation (\texttt{stack-3-blocks}), and dynamic bimanual manipulation
(\texttt{banana-hand-transfer}). Success is determined from the final scene state: whether the mug is grasped, the notebook is opened, the blocks are stacked, or the banana is transferred to the left hand. We additionally report DTW between the rollout and reference trajectories, separately measuring object position, object orientation, articulation, or wrist orientation as appropriate for each task.

\input{tables/ablations}

Table~\ref{tab:ablations} shows that removing hand trajectory optimization fails to learn across all tasks within the training budget. Replacing reference-centered residual actions with delta actions (where the policy predicts motion with respect to the current state, instead of the reference trajectory) succeeds in mug-lift and partially on the banana handover, but fails on notebook-opening and block-stacking tasks, and is less reliable in achieving the specified grasp change in mug task. This highlights the importance of the reference bias for tasks with long horizons or precise motion requirements. Restricting resets to the initial state also reduces performance, particularly
on the longer-horizon tasks, showing the benefit of sampling resets along the reference trajectory. Removing the interaction rewards preserves the mug lift success, but fail to achieve the grasp change, and performance also degrades on the notebook, block-stacking, and banana tasks. Overall, the results show that a reference containing both wrist and finger motion, together with reference-centered actions and trajectory-wide resets, is important for reliable learning.

Works such as \citet{de2026scaffolding} and \citet{chen2024objectcentricdexterousmanipulationhuman}  often provide a reference only for the wrist and learn finger actions from scratch. Building on these works, we test another variant of our method: the action space consists of wrist residuals and absolute finger-joint actions. 
We evaluate this setting on notebook opening and block stacking using wrist residuals and absolute finger
actions. It succeeds on notebook opening but fails on the longer-horizon block-stacking task, further demonstrating the value of including finger motion in the reference. 

\textbf{Sparse vs Dense Tracking.}
We study how densely interaction templates must be specified using \texttt{open-notebook} and \texttt{banana-handover}, two tasks with dense hand-object motion references from ARCTIC. From each dense reference, we construct three task specifications: templates at every frame, templates at a stride of six frames, and templates only at the start and end of the task. All
variants retain the same trajectory duration, with configurations between specified templates obtained through interpolation.

\begin{figure}[h!]
\centering
\revised{\includegraphics[width=0.84\linewidth,keepaspectratio]{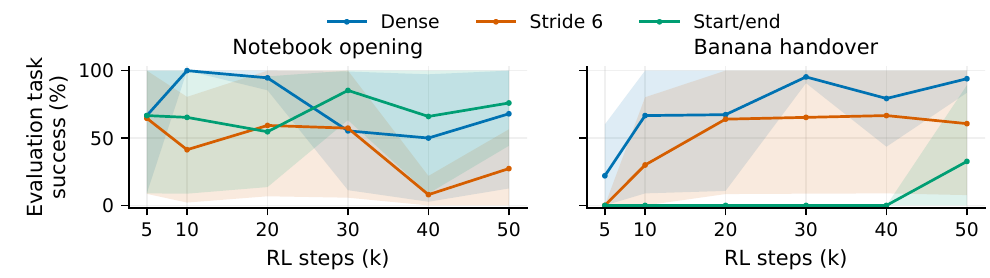}}
\vspace{-5pt}
\caption{Effect of template density on learning for open-notebook and banana-handover.}
\label{fig:sparsity_residual_ablations}
\vspace{-5pt}
\end{figure}

Figure~\ref{fig:sparsity_residual_ablations} shows that the right template density is task-dependent: open notebook remains stable with sparse waypoints, whereas banana handover benefits from denser temporal guidance. 
For tasks with discrete hand-object interaction phases (all tasks, except banana-handover), where the interaction type remains constant, a single source template per phase works well in practice.

\textbf{Dense Reference Tracking}
Finally, we test the framework's ability to work with dense task specifications using trajectories from Arctic MoCap data. We follow the evaluation setup used by \citet{mandi2025dexmachina}. Of the 7 task clips, five use matched object geometry, one is transferred to an in-category size variation, and one is transferred across categories from waffle-iron to a notebook.
\begin{figure}[h!]
    \centering
    \revised{\includegraphics[width=0.97\linewidth]{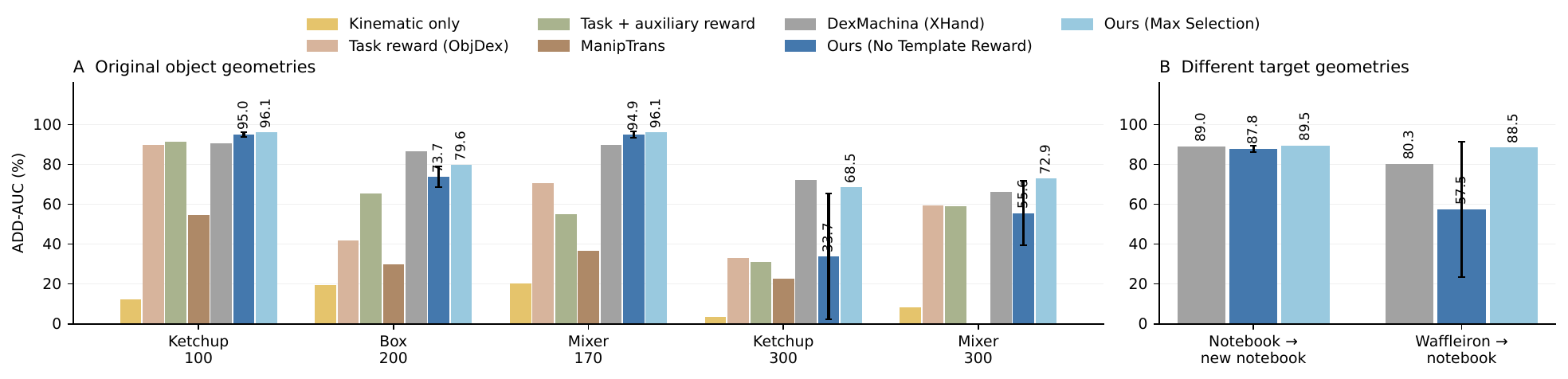}}
    \caption{Dense object tracking comparisons. The first five tasks use the original object geometry; the final two transfer the reference to an in-category or out-of-category target. For these transfer cases, the DexMachina values correspond to the original source and are included only as a reference. \texttt{Ours (Max Selection)} represents the best seed run (out of 3 seeds). \revised{Baseline values are reproduced from the published results of \cite{mandi2025dexmachina}.}}
    \label{fig:arctic}
\vspace*{-10pt}
\end{figure}

Our RL experiments use a different setting from previous experiments: we train with a maximum budget of 250k rollout steps and use the no-template reward variant of our method, because we find that template rewards can contradict highly dynamic object-motion targets, such as ARCTIC clips. Performance is evaluated using ADD-AUC, which measures object-trajectory tracking accuracy. 
  
Figure \ref{fig:arctic} compares our method with results from \cite{mandi2025dexmachina}. 
Except for two clips with high variance across seeds, \framework~achieves comparable dense-tracking performance while also supporting in-category and cross-category transfer, without requiring a hand-designed curriculum over object motion or gravity, as prior methods commonly do~\citep {mandi2025dexmachina, li2025maniptrans}.
  
\begin{figure}[!htbp]
\centering
\includegraphics[width=\linewidth,keepaspectratio]{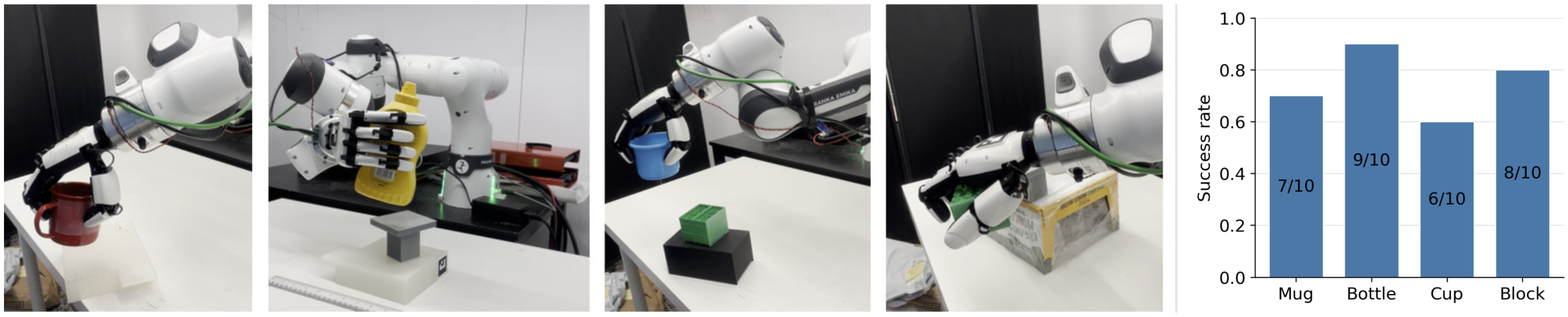}
\caption{{Sim-to-real grasp-and-lift evaluation. Left: successful rollouts with the mug and mustard bottle seen during simulation training and the unseen cup and block. Right: grasp success over 10 rollouts per object.}}
\label{fig:real_world_transfer_summary}
\vspace{-5pt}
\end{figure}

\textbf{Real World Deployment.} We deploy point-cloud policies trained entirely in simulation on four real objects: a mug and a mustard bottle from the training set, and two zero-shot transfers to different objects: a cup and a lego-block. We achieve object-level transfer through domain randomization during training. Figure~\ref{fig:real_world_transfer_summary} reports 10 grasp-and-lift trials per object. In this limited setting, the zero-shot policies successfully grasp and lift all four objects, including the two zero-shot transferred instances. This provides preliminary evidence of sim-to-real transfer of policies trained entirely in simulation.

\subsection{Generation cost and Multi-task RL}
\label{sec:multitask-rl-rebuttal}
Tasks can be authored from MoCap references, frames extracted from videos, or an agentic harness with human verification. In our authoring study, estimated manual effort for agentic authoring is $<1$ minute per task. Given a task specification, retargeting and RL run automatically without further manual intervention. Retargeting takes $\sim 10$ minutes on a CPU, while RL requires 1-20 hours of GPU time, depending on the task.

\begin{figure}[H]
\centering
\includegraphics[width=\dimexpr\linewidth-6pt\relax]{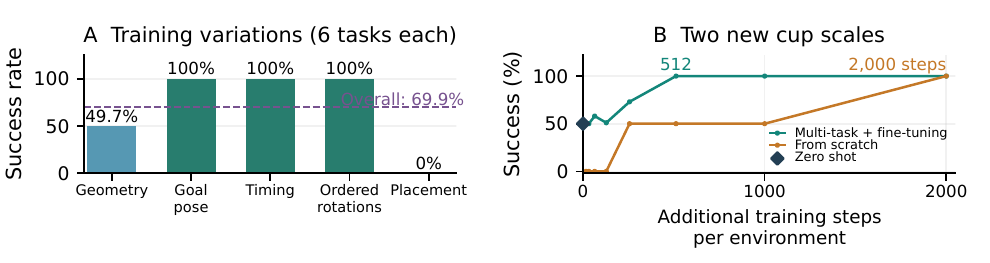}
\captionsetup{labelfont={color=revisioncolor},textfont={color=revisioncolor}}
\caption{(A) Shared-policy success on 30 training variants (six per category). (B) Zero-shot transfer and fine-tuning on two new cup scales, compared with training from scratch using the same architecture.}
\label{fig:multitask-rl-rebuttal}
\end{figure}

We show that multi-task RL policies can learn related task variations jointly and reduce the additional training needed for new variations. We adjust policy inputs and outputs to enable task-conditioned learning across 30 cup-manipulation variants (Figure~\ref{fig:multitask-rl-rebuttal}). The shared-policy results show $\sim$70\% success on the training task set and 50\% zero-shot success on two new cup scales. Fine-tuning on the new scales reaches comparable performance to training from scratch with 4x fewer training steps.

\section{Conclusion}
We presented \framework, a framework for synthesizing diverse dexterous manipulation data across objects and tasks, covering single-hand, bimanual, long-horizon, articulated, and dynamic tasks. Through systematic experimental analysis, we demonstrated the different capabilities of the framework: control over the generated behaviors along four axes, the reusability of interaction templates (used in specifying the desired behaviors) across object geometry variations, the ability to track both sparse and dense object reference motions, and a systematic ablation to emphasize the contribution of each component.

For task rollouts, please visit the \href{https://dex-link.github.io/}{project website}.

\section*{Limitations}

The framework has the following limitations. First, it requires an explicit task specification, which is less readily available than inputs such as videos. We believe current agentic systems are well-suited for this approach of specifying tasks. Second, the scope of the presented real-world experiments is limited to grasp-and-lift tasks. Finally, our experiments target only rigid objects. Sim-to-real transfer for a broader range of tasks and extending the approach to deformable objects are potential future work.

\bibliography{conference}
\bibliographystyle{conference}

\newpage
\appendix

\input{supp}

\end{document}

%% file: math_commands.tex
\usepackage{amsmath,amsfonts,bm}

\def\eqref#1{equation~\ref{#1}}

\def\1{\bm{1}}

\DeclareMathAlphabet{\mathsfit}{\encodingdefault}{\sfdefault}{m}{sl}
\SetMathAlphabet{\mathsfit}{bold}{\encodingdefault}{\sfdefault}{bx}{n}

%% file: notations.tex
\newcommand{\TaskSpec}{\mathcal{G}}
\newcommand{\SceneState}{\mathbf{s}}

\newcommand{\ObjMesh}{\mathcal{M}}

\newcommand{\HOTemplate}{\psi}

\newcommand{\SrcToTgtRotK}{R_k^{\mathrm{src}\rightarrow\mathrm{tgt}}}

\newcommand{\RefTraj}{\xi}

\newcommand{\cmark}{\checkmark}
\newcommand{\xmark}{\(\times\)}

\newcommand{\loss}{\mathcal{L}}
\newcommand{\SO}{\mathrm{SO}}
\newcommand{\SE}{\mathrm{SE}}
\newcommand{\FK}{\operatorname{FK}}
\newcommand{\wrist}{\mathbf{w}}
\newcommand{\Interp}{\operatorname{Interp}}

%% file: tables/template_transfer_tables.tex
\begingroup
\setlength{\tabcolsep}{4pt}
\renewcommand{\arraystretch}{1.15}
\begin{table*}[h!]
  \centering
    \caption{Template transfer to object geometry variations. Reported are mean success for lift tasks.}
    \vspace{-5pt}
  \label{tab:template-reuse}
  \begin{subtable}[t]{0.32\textwidth}
    \centering

    \resizebox{\linewidth}{!}{%
      \begin{tabular}{lcccc}
        \toprule
        & \multicolumn{4}{c}{Target-object scale} \\
        \cmidrule(lr){2-5}
        Hand-object interaction & \(0.70\) & \(0.85\) & \(1.00\) & \(1.25\) \\
        \midrule
        Bottle side & 1.0 & 1.0 & 1.0 & 0.0 \\
        Bottle top & 0.33 & 1.0 & 1.0 & 1.0 \\
        Mug handle & 0.67 & 1.0 & 0.67 & 0.0 \\
        Mug side & 0.99 & 1.0 & 1.0 & 1.0 \\
        Mug top & 1.0 & 1.0 & 1.0 & 0.67 \\
        \bottomrule
      \end{tabular}
    }
    \caption{\small Transfer to a target instance across scales.}
  \label{tab:scale-reuse}

  \end{subtable}%
  \hfill
  \begin{subtable}[t]{0.66\textwidth}
    \centering

    \resizebox{\linewidth}{!}{%
      \begin{tabular}{lllcc}
        \toprule
        Transfer type & Source category & Target category & Transferred (source $\rightarrow$ target) & Exact (target interaction)\\
        \midrule
        \multirow{3}{*}{\shortstack[l]{Within-category\\transfer}} & Mug & Mug & 1.0 & 1.0 \\
         & Bottle & Bottle & 1.0 & 1.0 \\
         & Trigger sprayer & Trigger sprayer & 1.0 & 1.0 \\
        \midrule
        \multirow{3}{*}{\shortstack[l]{Out-of-category\\transfer}} & Apple & Camera & 1.0 & 1.0 \\
         & Hammer & Wineglass & 1.0 & 0.67 \\
         & Apple & Mouse & 1.0 & 1.0 \\
        \bottomrule
      \end{tabular}
    }
        \caption{\footnotesize Instance- \& category-level transfer}
  \label{tab:across-category}
  \end{subtable}

\vspace{-10pt}

\end{table*}
\endgroup

%% file: tables/ablations.tex
\begin{table*}[t]
\centering
\caption{Ablations across four task categories. DTW is evaluated only on successful rollouts. }
\label{tab:ablations}
\vspace{-5pt}
\footnotesize
\setlength{\tabcolsep}{3.5pt}
\renewcommand{\arraystretch}{1.15}
\resizebox{\textwidth}{!}{%
\begin{tabular}{@{}l|ccc|cc|ccc|ccc@{}}
\toprule
& \multicolumn{3}{|c}{Mug grasp change} &
  \multicolumn{2}{|c}{Open notebook} &
  \multicolumn{3}{|c}{Stack blocks} &
  \multicolumn{3}{|c}{Banana handover} \\
\cmidrule(lr){2-4}
\cmidrule(lr){5-6}
\cmidrule(lr){7-9}
\cmidrule(l){10-12}
Method &
\shortstack{Lift success \\ (\%) $\uparrow$} &
\shortstack{Wrist rot. \\ DTW $\downarrow$} &
\revised{\shortstack{Ordered grasp \\ success (\%) $\uparrow$}} &
\shortstack{Success \\ (\%) $\uparrow$} &
\shortstack{Artic. \\ DTW $\downarrow$} &
\shortstack{Success \\ (\%) $\uparrow$} &
\shortstack{Pos. \\ DTW $\downarrow$} &
\shortstack{Rot. \\ DTW $\downarrow$} &
\shortstack{Success \\ (\%) $\uparrow$} &
\shortstack{Pos. \\ DTW $\downarrow$} &
\shortstack{Rot. \\ DTW $\downarrow$} \\
\midrule
\shortstack[l]{Obj. tracking \\+ wrist residual} &
\cellcolor{gray!20}{--} & \cellcolor{gray!20}{--} & \cellcolor{gray!20}{\revised{--}} & \cellcolor{green!25}{$100.0 \pm 0.0$} & \cellcolor{green!25}{$0.044 \pm 0.003$} & $0.0 \pm 0.0$ & -- & -- & \cellcolor{gray!20}{--} & \cellcolor{gray!20}{--} & \cellcolor{gray!20}{--} \\
\midrule
No reference traj &
-- & -- & \revised{$0.0 \pm 0.0$} & -- & -- & \revised{$0.0 \pm 0.0$} & -- & -- & -- & -- & -- \\
Delta actions &
\cellcolor{green!12}{$100.0 \pm 0.0$} & \cellcolor{green!12}{$0.210 \pm 0.042$} & \cellcolor{green!12}{\revised{$33.3 \pm 57.7$}} & $0.0 \pm 0.0$ & -- & $0.0 \pm 0.0$ & -- & -- & \cellcolor{green!12}{$66.7 \pm 57.7$} & \cellcolor{green!12}{$0.065 \pm 0.013$} & \cellcolor{green!12}{$0.421 \pm 0.041$} \\
Initial-only resets &
$100.0 \pm 0.0$ & $0.251 \pm 0.058$ & \revised{$33.3 \pm 57.7$} & $0.0 \pm 0.0$ & -- & \cellcolor{green!12}{\revised{$30.0 \pm 52.0$}} & \cellcolor{green!12}{\revised{$0.007$}} & \cellcolor{green!12}{\revised{$0.024$}} & $0.0 \pm 0.0$ & -- & -- \\
No interaction rewards &
$100.0 \pm 0.0$ & $0.314 \pm 0.023$ & \revised{$0.0 \pm 0.0$} & $66.7 \pm 57.7$ & $0.042 \pm 0.001$ & $0.0 \pm 0.0$ & -- & -- & $33.3 \pm 57.7$ & $0.056$ & $0.483$ \\
\textbf{Full method} &
\cellcolor{green!25}{$100.0 \pm 0.0$} & \cellcolor{green!25}{$0.163 \pm 0.091$} & \cellcolor{green!25}{\revised{$92.0 \pm 12.2$}} & \cellcolor{green!12}{$100.0 \pm 0.0$} & \cellcolor{green!12}{$0.055 \pm 0.022$} & \cellcolor{green!25}{\revised{$86.7 \pm 16.2$}} & \cellcolor{green!25}{\revised{$0.008 \pm 0.002$}} & \cellcolor{green!25}{\revised{$0.035 \pm 0.018$}} & \cellcolor{green!25}{$97.3 \pm 4.6$} & \cellcolor{green!25}{$0.037 \pm 0.002$} & \cellcolor{green!25}{$0.371 \pm 0.008$} \\
\bottomrule
\end{tabular}%
}
\vspace{-10pt}
\end{table*}

%% file: supp.tex
\renewcommand{\check}[1]{{\color{blue}\footnotesize[CHECK: #1]}}
\newcommand{\suppimp}[1]{{\color{red}\footnotesize[suppimg: #1]}}

\section{Reference Trajectory Construction Details}

\subsection{Retargeting Objective}

\begin{figure*}[h!]
    \centering
    \includegraphics[width=0.86\textwidth]{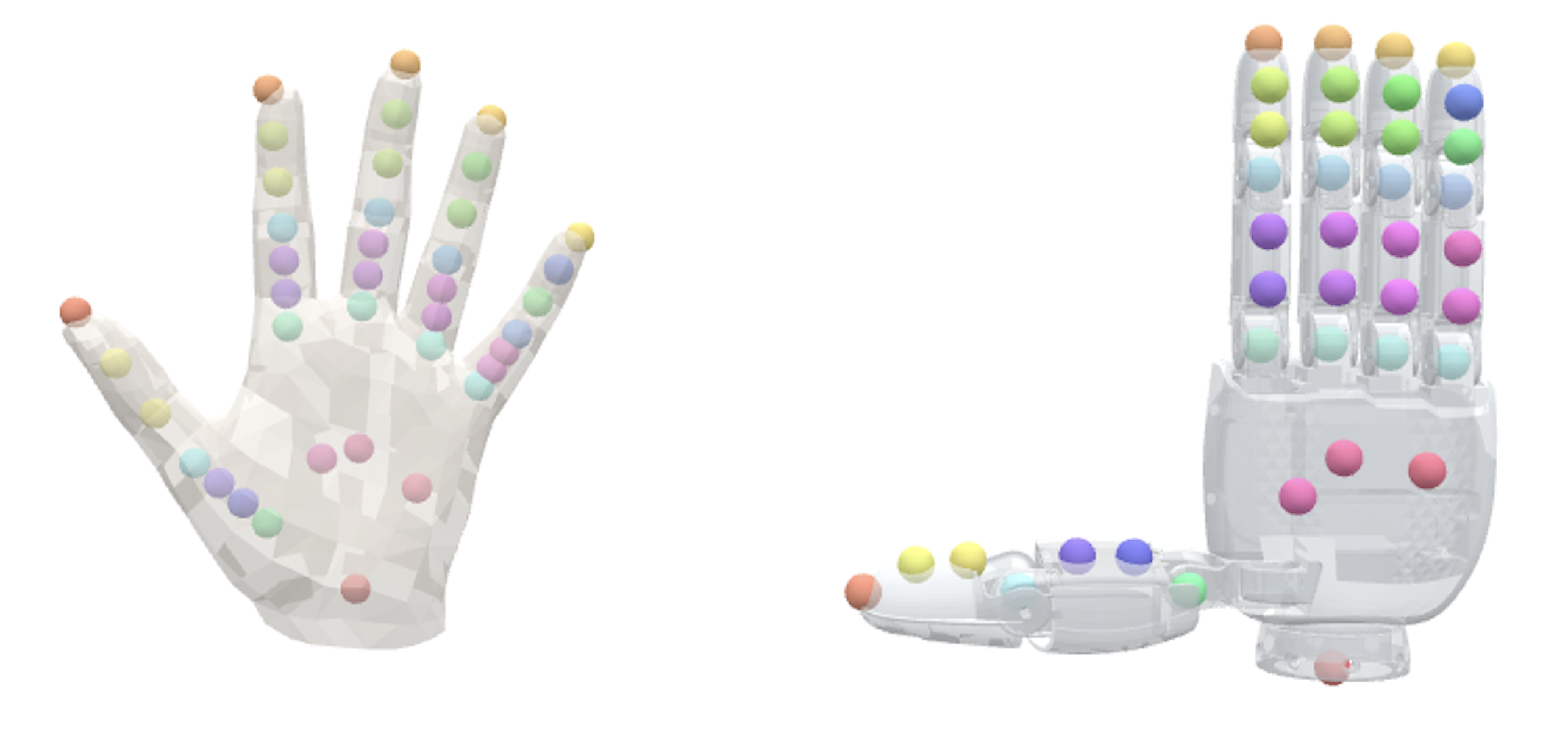}
    \caption{Hand keypoints used by the retargeting and reward terms, with
    corresponding points on the human and robot hands.}
    \label{fig:supp_hand_kpts}
\end{figure*}

For interaction template \(\HOTemplate_k\), the robot keypoints in the
target-object frame are
\(P_k^{\mathrm r}=\wrist_k\!\cdot\!\FK(q_k)\), and the aligned human
keypoints are \(\bar P_k^{\mathrm h}\). Retargeting compares these keypoints
using the hand-configuration loss \(\loss_{\mathrm{cfg}}\) and compares their
relationships to the corresponding objects using
\(\loss_{\mathrm{rel}}\). Figure~\ref{fig:supp_hand_kpts} shows the shared
human and robot keypoints.

\paragraph{Hand-configuration similarity.}
Table~\ref{tab:handcfg_relations} defines the pair and triplet relations used by
\(\loss_{\mathrm{cfg}}\).

\begin{table}[H]
\caption{Hand keypoint relations used in \(\loss_{\mathrm{cfg}}\). The last
column indicates whether the vector is normalized before comparison.}
\centering
\scriptsize
\setlength{\tabcolsep}{4pt}
\begin{tabular}{p{0.23\columnwidth}p{0.58\columnwidth}c}
\toprule
Group & Relation set & Norm. \\
\midrule
Wrist$\rightarrow$tip &
\(\{\,\texttt{wrist}\!\rightarrow\!f_{\texttt{tip}} \mid
f\in\{\texttt{thumb,index,middle,ring,little}\}\,\}\) & Yes \\
Base$\rightarrow$tip &
\(\{\,f_{\texttt{base}}\!\rightarrow\!f_{\texttt{tip}} \mid
f\in\{\texttt{thumb,index,middle,ring,little}\}\,\}\) & Yes \\
Thumb-tip opposition &
\(\{\,\texttt{thumb\_tip}\!\rightarrow\!f_{\texttt{tip}} \mid
f\in\{\texttt{index,middle,ring,little}\}\,\}\) & No \\
Palm orientation &
\((\texttt{thumb\_base}-\texttt{wrist}) \times
(\texttt{little\_base}-\texttt{wrist})\) & Yes \\
\bottomrule
\end{tabular}
\label{tab:handcfg_relations}
\end{table}
For \(s\in\{\mathrm r,\mathrm h\}\), let \(p^s(u)\) denote the position of
keypoint \(u\) in \(P^{\mathrm r}\) or \(P^{\mathrm h}\). For a pair
\((u,v)\) in Table~\ref{tab:handcfg_relations}, define
\[
v_{uv}^s=p^s(v)-p^s(u), \qquad
\hat v_{uv}^s=v_{uv}^s/(\|v_{uv}^s\|_2+\epsilon).
\]
For the palm triplet \((u,v,w)\), let
\[
c_{uvw}^s=(p^s(v)-p^s(u))\times(p^s(w)-p^s(u)), \qquad
\hat c_{uvw}^s=c_{uvw}^s/(\|c_{uvw}^s\|_2+\epsilon).
\]
The loss compares the corresponding relations of the two hands:
\[
\loss_{\mathrm{cfg}}(P^{\mathrm r},P^{\mathrm h})
=
\frac{1}{|\mathcal{P}|}\sum_{(u,v)\in\mathcal{P}}
d_{\mathrm{pair}}(u,v)
+
\frac{1}{|\mathcal{T}|}\sum_{(u,v,w)\in\mathcal{T}}
d_{\mathrm{triplet}}(u,v,w),
\]
where normalized relations use
\(2(1-\hat v^{\mathrm r}\!\cdot\!\hat v^{\mathrm h})\) or
\(2(1-\hat c^{\mathrm r}\!\cdot\!\hat c^{\mathrm h})\), and unnormalized relations use squared
Euclidean distance.

\paragraph{Hand-object relational similarity.}
The hand-object term uses a differentiable BPS-based descriptor~\citep{Prokudin2019EfficientLO}.
For each object, the optimizer samples a point cloud with surface normals and
normalized object-coordinate features. For a hand keypoint \(b_i\) (marked in
Figure~\ref{fig:supp_hand_kpts}) and object
surface samples \(x_j\), we avoid a hard nearest-neighbor assignment. Instead,
we use exponential soft-nearest weights
\[
a_{ij}
=
\frac{\exp(-\beta\|b_i-x_j\|_2^2)}
{\sum_k \exp(-\beta\|b_i-x_k\|_2^2)}.
\]
The descriptor then uses the expected closest point, normal, and feature
channels under these weights:
\[
\bar x_i=\sum_j a_{ij}x_j,\qquad
\bar n_i=\frac{\sum_j a_{ij}n_j}{\|\sum_j a_{ij}n_j\|_2},\qquad
\bar f_i=\sum_j a_{ij}f_j.
\]
This exponential weighting makes the hand-object descriptor differentiable
with respect to the hand keypoint positions, so the resulting objective can be
optimized by SNOPT. The descriptor stores the displacement to the soft-nearest
surface point, its normal, and its normalized object-coordinate feature:
\[
\Psi_{\mathrm{rel}}(P,\ObjMesh) = [D, N, F].
\]
Using the notation from the main paper, we compute
\[
\begin{aligned}
(D^{\mathrm r},N^{\mathrm r},F^{\mathrm r})
  &=\Psi_{\mathrm{rel}}(P_k^{\mathrm r},\ObjMesh_k^{\mathrm{tgt}}),\\
(D^{\mathrm h},N^{\mathrm h},F^{\mathrm h})
  &=\Psi_{\mathrm{rel}}(\bar P_k^{\mathrm h},\bar{\ObjMesh}_k^{\mathrm{src}}).
\end{aligned}
\]

We compare the robot and source BPS descriptors using four terms. Let
\(d_{r,i}^2=\|D_i^{\mathrm r}\|_2^2\) and
\(d_{h,i}^2=\|D_i^{\mathrm h}\|_2^2\). The distance term matches the
relative hand-object distance by
\[
\loss_{\mathrm{dist}}
=
\sum_i\left|1-\frac{d_{r,i}^2}{d_{h,i}^2}\right|.
\]
The direction term compares normalized displacement directions,
\[
\loss_{\mathrm{dir}}
=
\sum_i\left(1-\hat D_i^{\mathrm r}\cdot \hat D_i^{\mathrm h}\right).
\]
The normal term compares object normals, emphasizing near-contact relations,
\[
\loss_{\mathrm{normal}}
=
\sum_i
\left(1-N_i^{\mathrm r}\cdot N_i^{\mathrm h}\right)
\exp\!\left(-\frac{d_{h,i}^2}{2(0.05)^2}\right).
\]
Finally, when feature channels are present, the feature term is
\[
\loss_{\mathrm{feat}}=\sum_i |F_i^{\mathrm r}-F_i^{\mathrm h}|.
\]
The hand-object relational loss is
\[
\loss_{\mathrm{rel}}
=
\loss_{\mathrm{dist}}
+\loss_{\mathrm{dir}}
+\loss_{\mathrm{normal}}
+\loss_{\mathrm{feat}}.
\]

\paragraph{Final Objective.}
For each interaction template, the retargeting objective is
\[
    \loss_{\mathrm{ret}}
    =
    \loss_{\mathrm{cfg}}
    +\lambda\loss_{\mathrm{rel}},
\]
matching the objective in the main paper. 

\paragraph{Retargeting hyperparameters.}
Table~\ref{tab:supp_retarget_hyperparams} summarizes the main numerical
settings used for object-aware retargeting. The hand-object descriptor is
computed by sampling the target object surface and evaluating a soft nearest
surface point for each selected hand keypoint.

\begin{table}[h]
\centering
\small
\caption{Retargeting hyperparameters for XHand and the Drake/SNOPT solver.}
\label{tab:supp_retarget_hyperparams}
\begin{tabular}{@{}p{0.3\linewidth}p{0.66\linewidth}@{}}
\toprule
\textbf{Setting} & \textbf{Value} \\
\midrule
Solver tolerance & \(10^{-4}\) \\
Object surface samples & \(512\) \\
Hand keypoints in BPS descriptor & \(39\) \\
BPS channels per keypoint & \(9\): 3D displacement, surface normal, normalized object coordinates \\
Flattened BPS descriptor size & \(39 \times 9 = 351\) \\
Relational-loss weight \(\lambda\) & \(10^{-3}\) \\
\bottomrule
\end{tabular}
\end{table}

\begin{figure*}[t]
    \centering
    \includegraphics[width=0.92\textwidth]{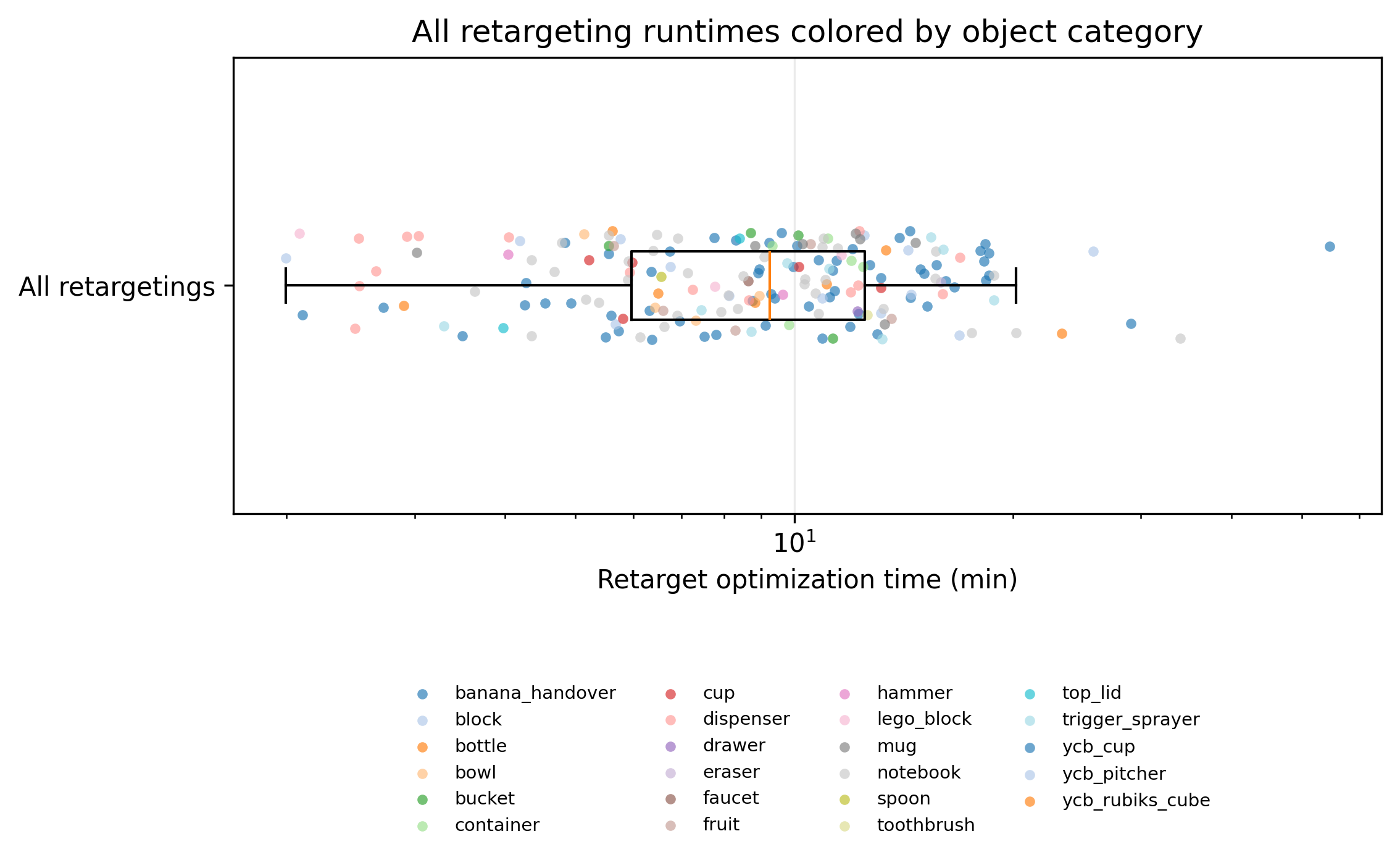}
    \caption{Retargeting runtime by object category. This cost is paid once per
    task/template before RL training and deployment.}
    \label{fig:supp_retarget_timing}
\end{figure*}

Figure~\ref{fig:supp_retarget_timing} summarizes the retargeting
runtime across object categories. On average, optimization takes
about \(10\) minutes on a single CPU. Waypoints are retargeted independently and
can be parallelized.

\subsection{Resolving Penetration and Enforcing Contacts}

The optimization produces coarse retargeting results that can contain small
hand-object penetrations. We therefore apply a lightweight post-processing step
that performs sequential line search over finger-joint groups. The line search
selects nearby poses that remove robot-object penetration and, when the source
template indicates contact at a keypoint, moves that keypoint closer to the
object. This step prevents simulation rollouts from starting with
interpenetrating geometry.

\section{RL Refinement Details}

\begin{table}[H]
    \centering
    \small
    \caption{Policy variants and control setup.}
    \label{tab:supp-policy-variants}
    \begin{tabular}{@{}p{0.30\linewidth}p{0.64\linewidth}@{}}
        \toprule
        \textbf{Setting} & \textbf{Value} \\
        \midrule
        Simulation robot & Floating XHand with 6-DoF wrist \\
        Real robot & Franka arm with XHand \\
        Simulation policy & State based \\
        Real-world policy & Point-cloud based; no privileged simulator state \\
        Controller & Position control with targets interpolated over 8 physics steps \\
        \bottomrule
    \end{tabular}
\end{table}

Tables~\ref{tab:supp_policy_obs_dims} and~\ref{tab:supp_policy_action_dims}
specify the single-hand state-policy interface.

\begin{table}[H]
\centering
\small
\caption{Policy observation dimensions for the single-hand state-based experiments.}
\label{tab:supp_policy_obs_dims}
\begin{tabular}{@{}lr@{}}
\toprule
Observation & Dim. \\
\midrule
\textbf{Robot state} & 69 \\
\quad Hand joint positions & 12 \\
\quad Wrist position and orientation \(R_x,R_y\) & 9 \\
\quad Hand link positions & 48 \\
\midrule
\textbf{Reference state} & 21 \\
\quad Reference wrist position & 3 \\
\quad Reference wrist orientation \(R_x,R_y\) & 6 \\
\quad Reference finger joint positions & 12 \\
\midrule
\textbf{Object state} & 10 \\
\quad Object position & 3 \\
\quad Object orientation \(R_x,R_y\) & 6 \\
\quad Object/articulated-part DoFs & 1 \\
\midrule
\textbf{Task progress}: reference timestep & 1 \\
\midrule
\textbf{Contact state}: tactile-link signals & 15 \\
\midrule
Total & 116 \\
\bottomrule
\end{tabular}
\end{table}

\begin{table}[t]
\centering
\small
\caption{Policy action dimensions for the single-hand state-based experiments.}
\label{tab:supp_policy_action_dims}
\begin{tabular}{@{}lr@{}}
\toprule
Action & Dim. \\
\midrule
\textbf{Robot hand action} & 18 \\
\quad Wrist Cartesian translation delta & 3 \\
\quad Wrist rotation-vector delta & 3 \\
\quad Controlled hand-joint deltas & 12 \\
\midrule
Total & 18 \\
\bottomrule
\end{tabular}
\end{table}

\subsection{Policy Architecture}

\begin{table}[H]
    \centering
    \small
    \caption{Policy architecture.}
    \label{tab:supp-policy-architecture}
    \begin{tabular}{@{}p{0.28\linewidth}p{0.66\linewidth}@{}}
        \toprule
        \textbf{Component} & \textbf{Specification} \\
        \midrule
        Actor/critic & Separate networks; observation-group embeddings are summed and layer-normalized before each MLP \\
        Actor output & Gaussian residual policy \\
        Smoothed target & Correct $q_t$ from reference by $\Delta q_t$ and blend neighboring corrections with a Gaussian temporal window \\
        Network sizes & Actor: approximately 200k parameters; critic: approximately 25k \\
        Point-cloud policy & PointNet-style hand--object encoder plus non-point-cloud state; state-based critic \\
        \bottomrule
    \end{tabular}
\end{table}

\subsection{Reward Design}
\label{sec:supp_reward_design}

The total reward consists of the terms shown in
Table~\ref{tab:supp_reward_weights}. All dense reward values lie in
\([0,1]\), and the dense reward weights are normalized to sum to \(1.0\). The
task-completion reward is added separately.

\begin{table}[t]
\centering
\small
\caption{Reward weights: Dense reward weights are
normalized to one; the completion reward is
added as a separate sparse bonus.}
\label{tab:supp_reward_weights}
\begin{tabular}{@{}lcc@{}}
\toprule
Reward term & Unnorm. weight & Norm. weight \\
\midrule
\textbf{Dense reward} & 1.20 & 1.000 \\
\quad Object tracking & 0.75 & 0.625 \\
\quad Hand-configuration similarity & 0.10 & 0.083 \\
\quad Hand-object relational similarity & 0.10 & 0.083 \\
\quad Retargeted joint-pose closeness & 0.20 & 0.167 \\
\quad Joint smoothness & 0.05 & 0.042 \\
\midrule
\textbf{Completion reward} & 10.00 & 10.00 \\
\bottomrule
\end{tabular}
\end{table}

\subsection{Environment Reset}

The reset sampler initializes robot and object state from a reference timestep;
sampling probability is proportional to the negative exponential of its running
visitation density.

\subsection{Ablations}
Table \ref{tab:ablation_variants} describe the ablation variants considered in Table \ref{tab:ablations}, with respect to change in the RL training.

\providecommand{\cmark}{\ensuremath{\checkmark}}
\providecommand{\xmark}{\ensuremath{\times}}

\begin{table*}[t]
    \centering
    \caption{Components retained in each method variant. A checkmark denotes an enabled component.}
    \label{tab:ablation_variants}
    \setlength{\tabcolsep}{3.5pt}
    \renewcommand{\arraystretch}{1.12}
    \resizebox{\textwidth}{!}{%
        \begin{tabular}{@{}lcccccccc@{}}
            \toprule
            & \multicolumn{4}{c}{\textbf{Reference and policy}} &
              \multicolumn{4}{c}{\textbf{Interaction rewards}} \\
            \cmidrule(lr){2-5}\cmidrule(l){6-9}
            \textbf{Variant} &
            \shortstack{Optimized robot\\hand reference} &
            \shortstack{Trajectory-\\conditioned actor} &
            \shortstack{Reference-relative\\residual actions} &
            \shortstack{All-phase\\resets} &
            \shortstack{Hand\\configuration} &
            \shortstack{Hand--object\\BPS} &
            \shortstack{Retargeted-joint\\tracking} &
            \shortstack{Wrist-velocity\\tracking} \\
            \midrule
            Full                     & \cmark & \cmark & \cmark & \cmark & \cmark & \cmark & \cmark & \cmark \\
            No hand trajopt          & \xmark & \xmark & \xmark & \xmark & \cmark$^{*}$ & \cmark$^{*}$ & \xmark & \xmark \\
            Delta actions            & \cmark & \cmark & \xmark & \cmark & \cmark & \cmark & \cmark & \cmark \\
            Initial-only resets      & \cmark & \cmark & \cmark & \xmark & \cmark & \cmark & \cmark & \cmark \\
            No interaction rewards   & \cmark & \cmark & \cmark & \cmark & \xmark & \xmark & \xmark & \xmark \\
            \bottomrule
        \end{tabular}%
    }
    \vspace{2pt}

    \parbox{\textwidth}{\scriptsize
        \textbf{Variant details.}
        Delta actions predict joint-position changes from the current hand state.
        Initial-only resets start every episode from the first task state rather than from states spanning the reference trajectory.
        $^{*}$Without hand trajopt, the hand-configuration and BPS terms are evaluated against the human templates; robot-reference observations and robot-reference-dependent rewards are removed.
    }
\end{table*}

\subsection{PPO and Simulator Hyperparameters}

Tables~\ref{tab:supp_sim_hyperparams} and~\ref{tab:supp_ppo_hyperparams}
summarize the simulator and optimization settings.

\begin{table}[H]
\centering
\small
\caption{Simulation hyperparameters.}
\label{tab:supp_sim_hyperparams}
\begin{tabular}{@{}lr@{}}
\toprule
\textbf{Setting} & \textbf{Value} \\
\midrule
Simulation timestep & \(1/120\) s \\
Control decimation & \(8\) \\
Gravity & \(-9.81\,\mathrm{m/s^2}\) \\
Default object density & \(1200\,\mathrm{kg/m^3}\) \\
Object-mass randomization & None \\
\bottomrule
\end{tabular}
\end{table}

\begin{table}[H]
\centering
\small
\caption{PPO hyperparameters.}
\label{tab:supp_ppo_hyperparams}
\begin{tabular}{@{}lr@{}}
\toprule
\textbf{Hyperparameter} & \textbf{Value} \\
\midrule
Optimizer & AdamW \\
Policy learning rate & \(10^{-4}\) \\
Value learning rate & \(10^{-4}\) \\
Discount factor \((\gamma)\) & \(0.95\) \\
GAE parameter \((\lambda)\) & \(0.95\) \\
Rollout horizon & \(8\) \\
Minibatch size & \(1024\) \\
Update epochs & \(4\) \\
Clip ratio \((\epsilon)\) & \(0.2\) \\
Entropy coefficient & \(10^{-3}\) \\
Value-loss coefficient & \(0.5\) \\
Advantage normalization & No \\
\bottomrule
\end{tabular}
\end{table}

\section{DTW Metric Details}
\label{sec:supp-dtw}

We compute dynamic time warping with the \texttt{dtw-python} library, using a
normalized Sakoe--Chiba window of \(0.10\) times the sequence length. We use
Euclidean local costs for positions, articulation values, and finger joints,
and quaternion geodesic distance for orientations; reported values are the
library-normalized distances.

\clearpage
\section{Task Details}
\label{sec:suppv2-task-details}

We use $H$ for the number of human-sourced interaction-frame instances, $K$
for the number of explicit robot keyframes after adding geometrically derived
states, and $T$ for the compiled reference horizon. Thus, $H/T$ measures direct
human-template coverage, while $(T-K)/T$ is the fraction generated by reach,
interpolation, and hold phases. 

\subsection{Reusability}
\label{sec:suppv2-task-reuse}

\begin{figure*}[t]
    \centering
    \setlength{\tabcolsep}{3pt}
    \begin{tabular}{@{}ccccc@{}}
        \textbf{Bottle side} & \textbf{Bottle top} & \textbf{Mug handle} & \textbf{Mug side} & \textbf{Mug top} \\
        \includegraphics[width=0.15\textwidth]{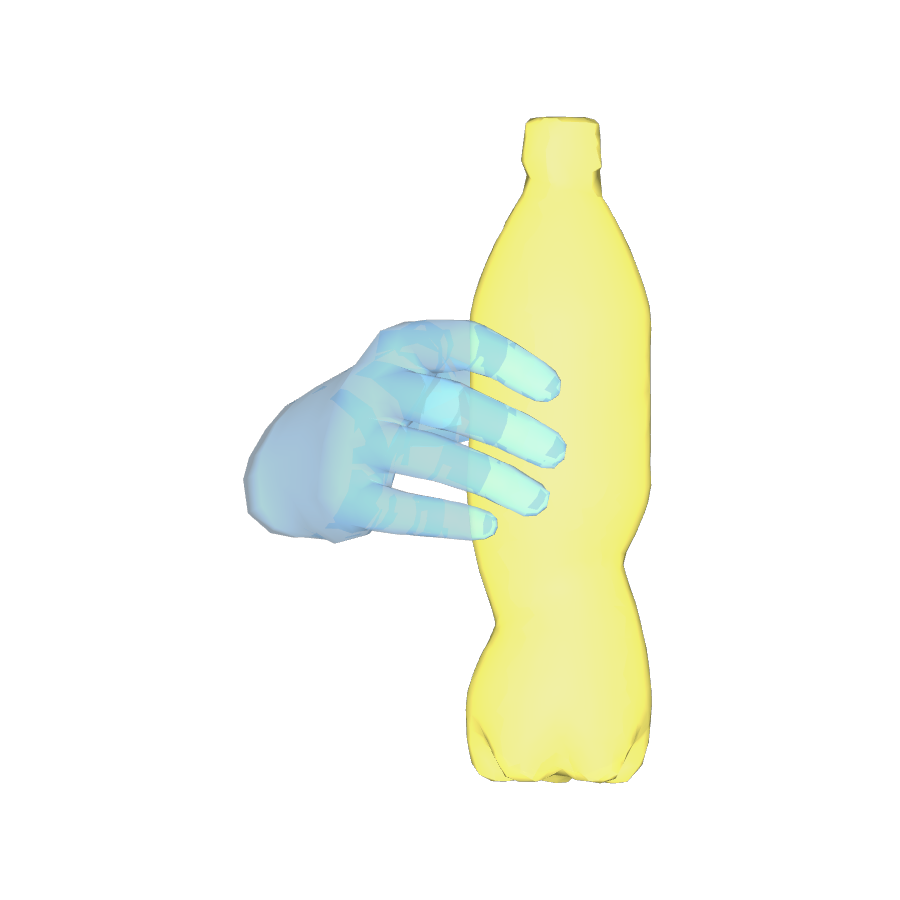} &
        \includegraphics[width=0.15\textwidth]{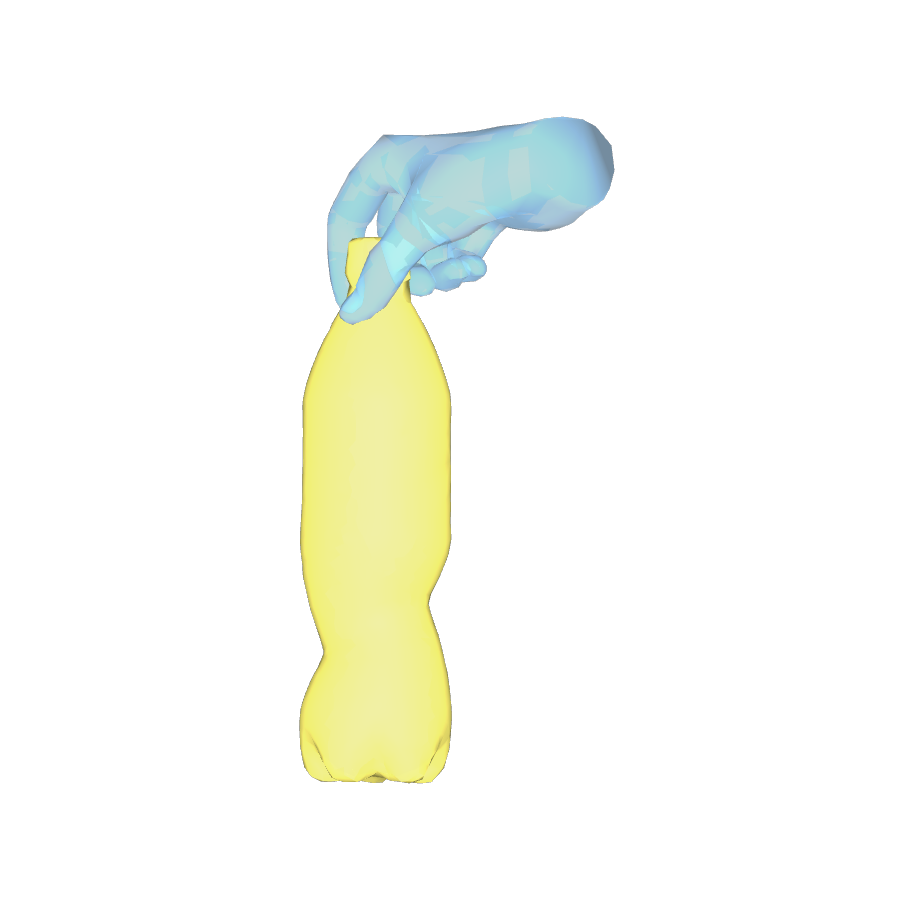} &
        \includegraphics[width=0.15\textwidth]{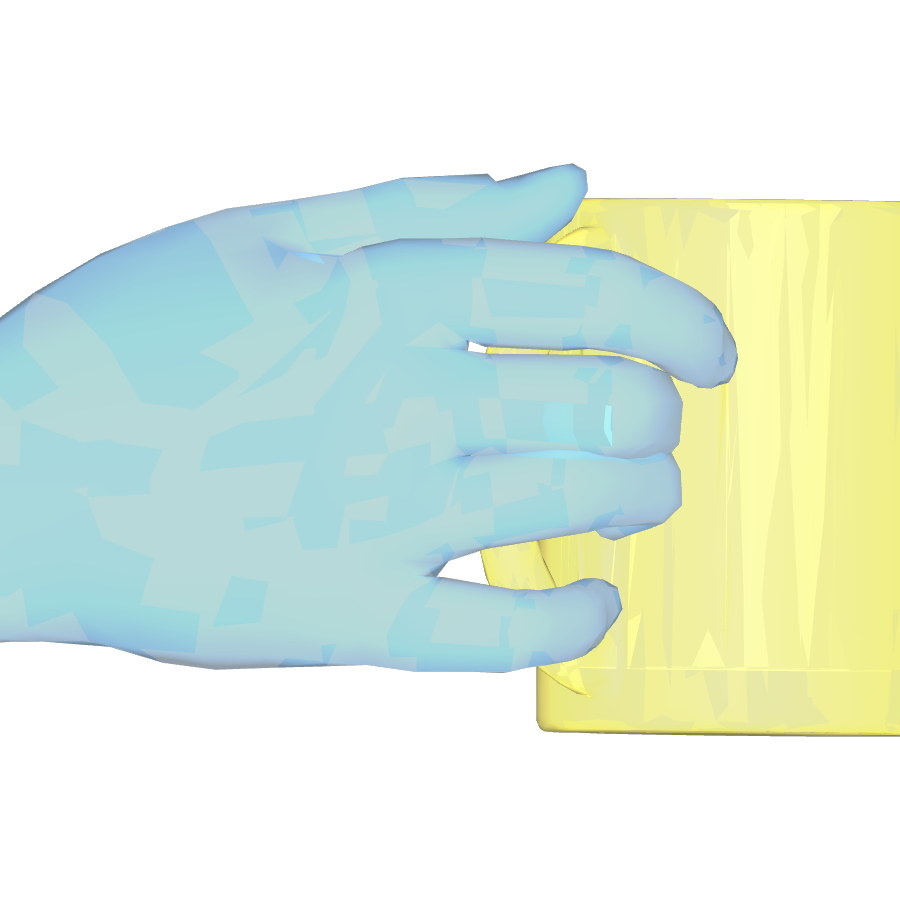} &
        \includegraphics[width=0.15\textwidth]{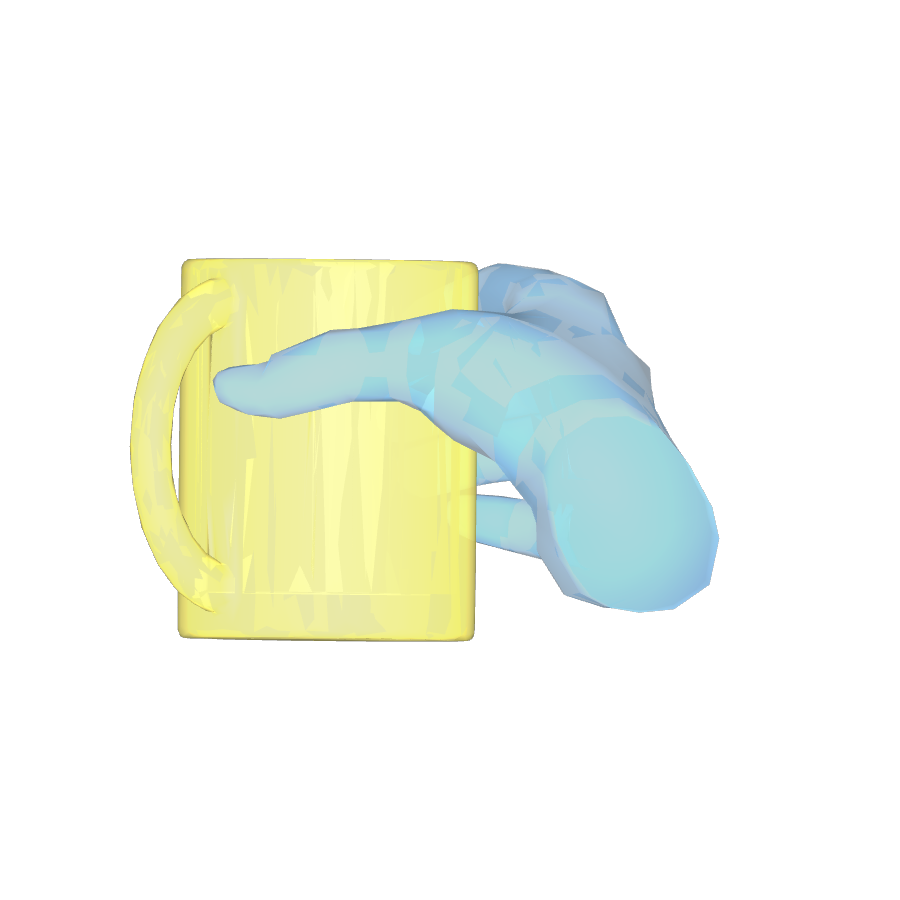} &
        \includegraphics[width=0.15\textwidth]{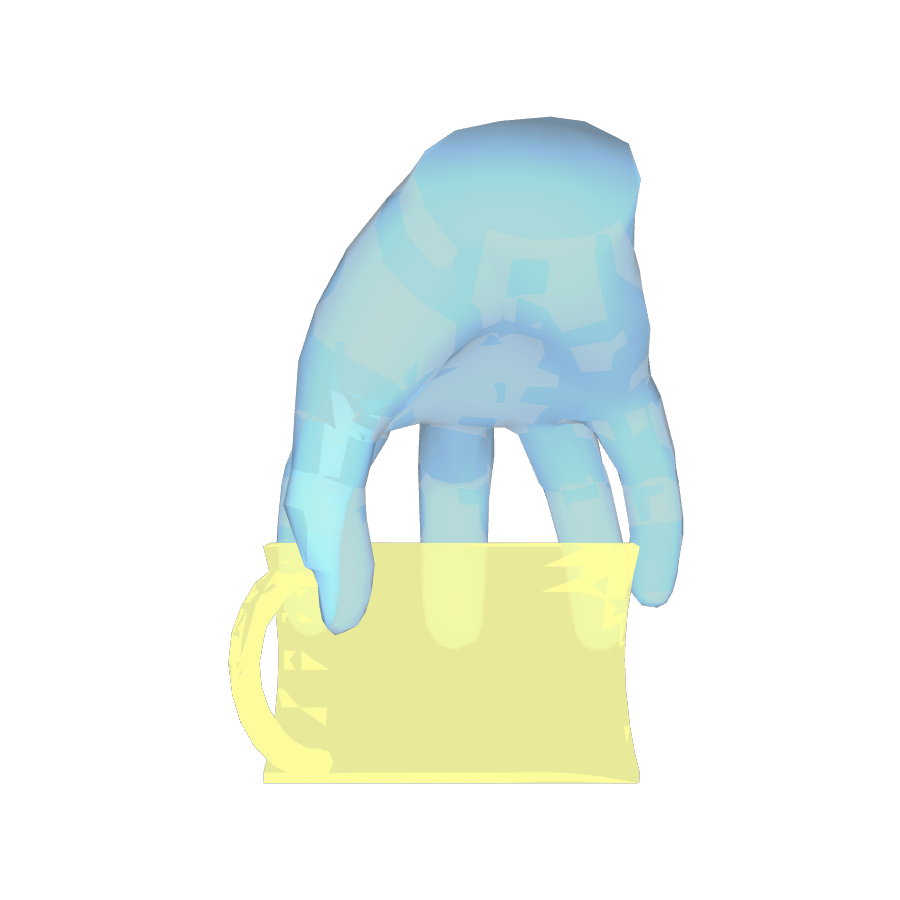} \\
        \multicolumn{5}{c}{\footnotesize Human source interactions} \\[2pt]
        \includegraphics[width=0.18\textwidth]{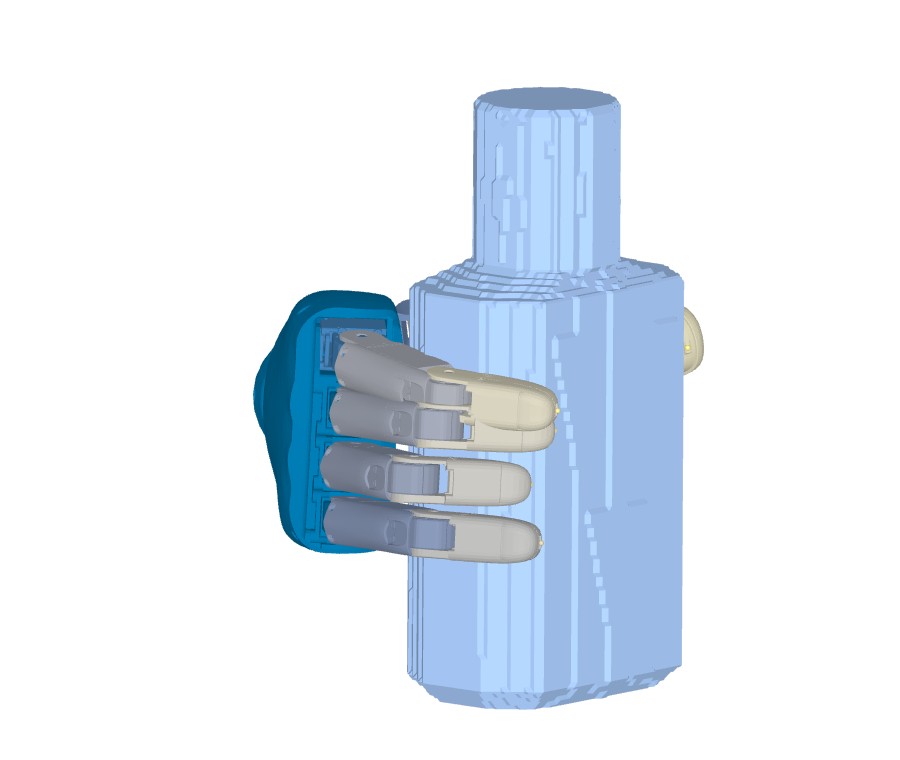} &
        \includegraphics[width=0.18\textwidth]{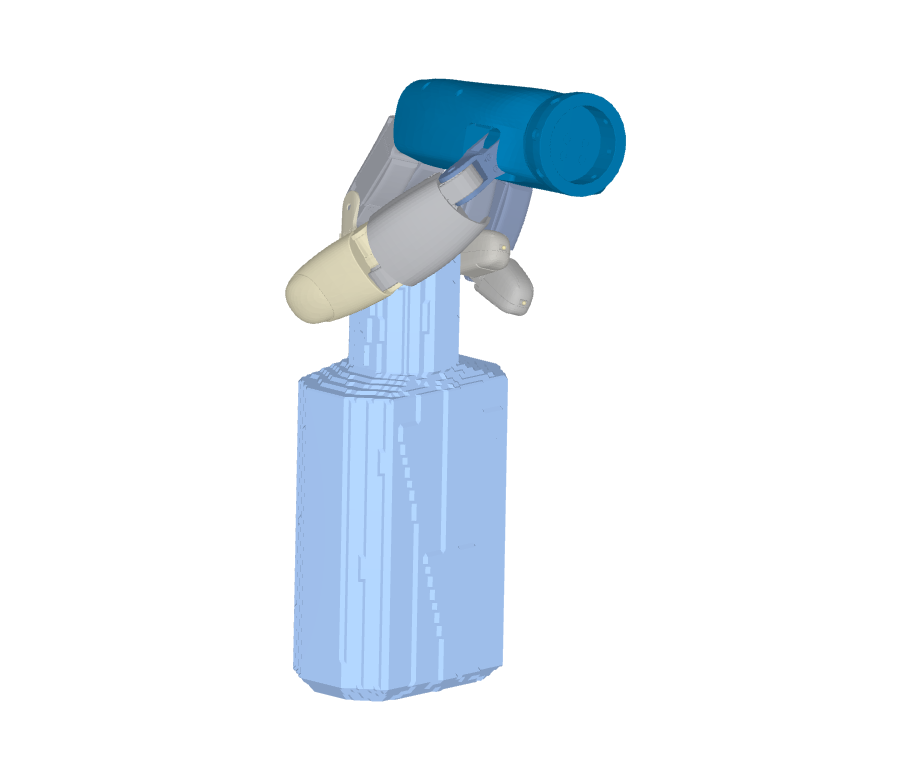} &
        \includegraphics[width=0.18\textwidth]{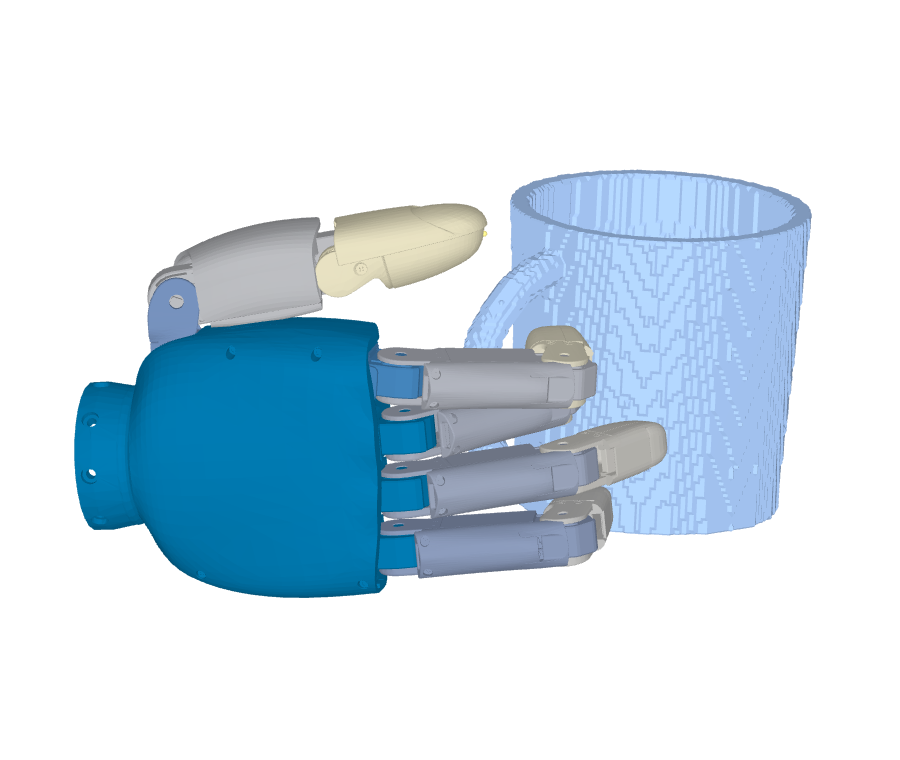} &
        \includegraphics[width=0.18\textwidth]{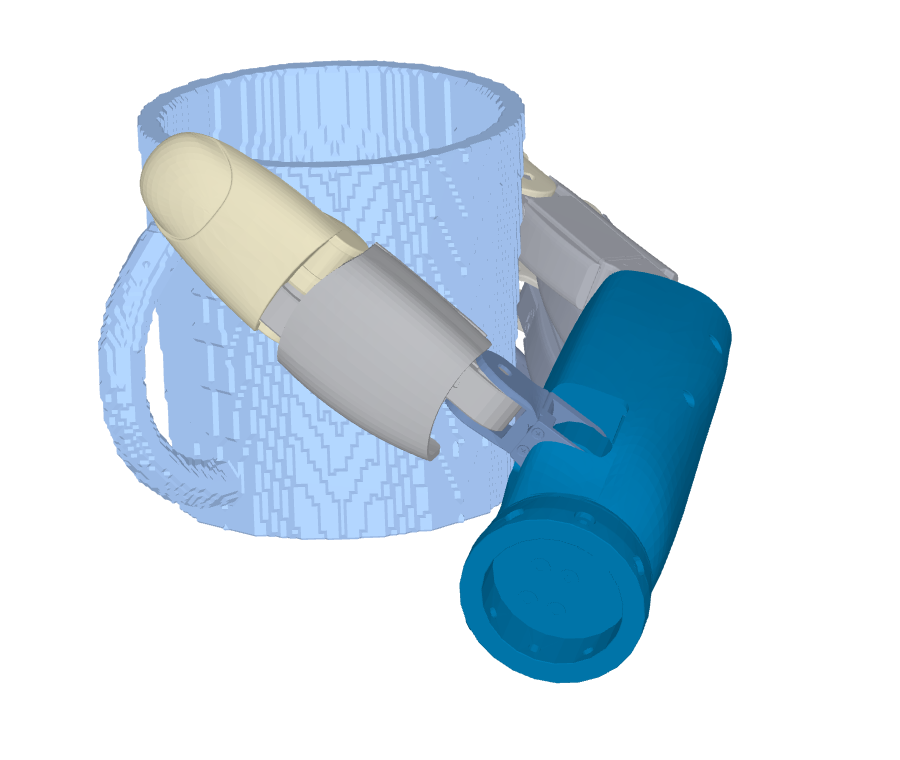} &
        \includegraphics[width=0.18\textwidth]{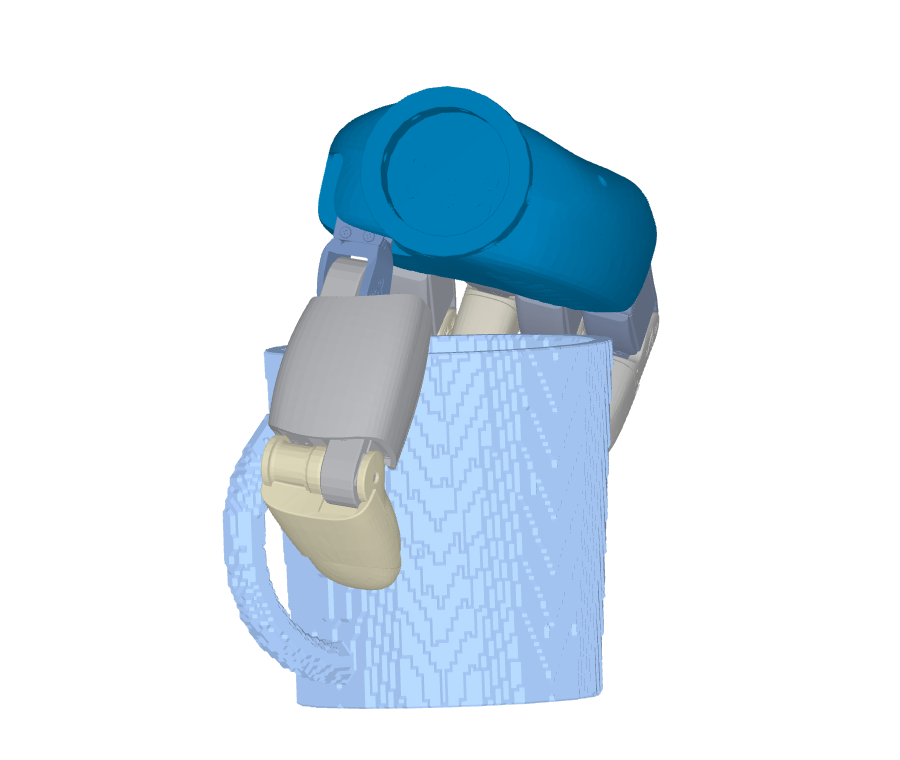} \\
        \multicolumn{5}{c}{\footnotesize Retargeted robot interactions at $1.00\times$} \\
    \end{tabular}
    \caption{Source and target interactions for scale reuse. Within each row,
    the same source template is used at all four target-object scales. The
    $1.00\times$ target is shown for reference.}
    \label{fig:suppv2-scale-template-grid}
\end{figure*}

\begin{figure*}[t]
    \centering
    \scriptsize
    \setlength{\tabcolsep}{3pt}
    \begin{tabular}{@{}lcccc@{}}
        & \multicolumn{2}{c}{\textbf{Transferred template}} & \multicolumn{2}{c}{\textbf{Exact-target template}} \\
        \textbf{Pair} & \textbf{Source} & \textbf{Robot target} & \textbf{Source} & \textbf{Robot target} \\
        Mug top &
        \includegraphics[width=0.13\textwidth]{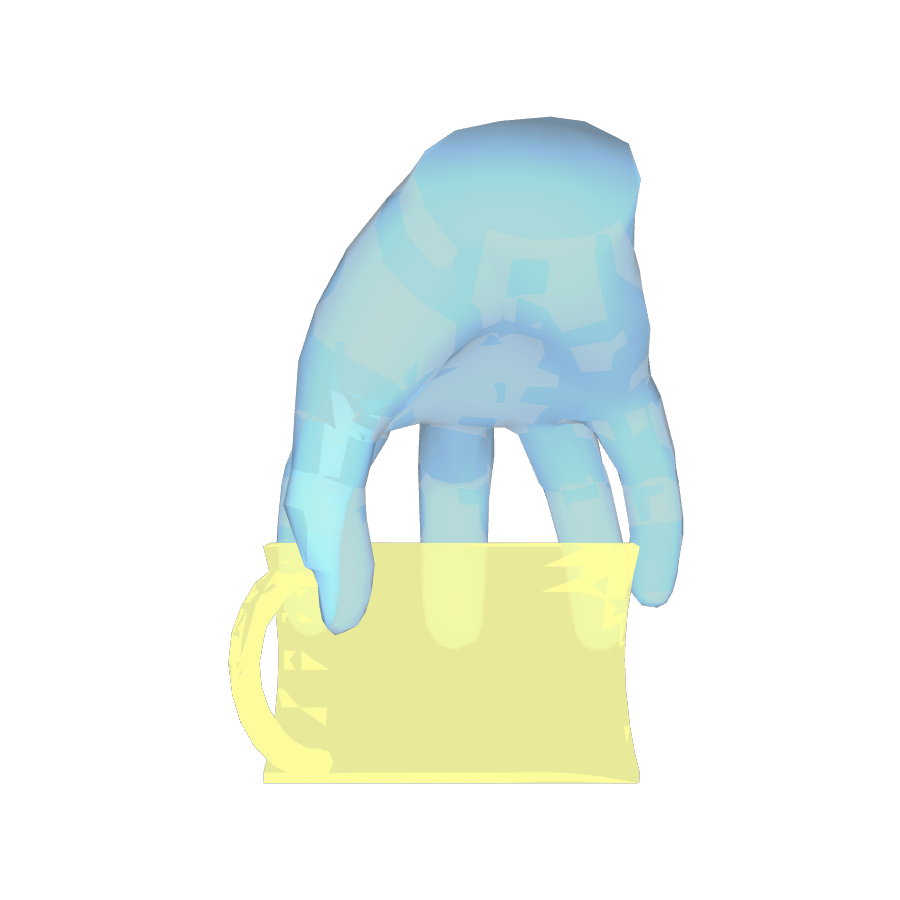} &
        \includegraphics[width=0.18\textwidth]{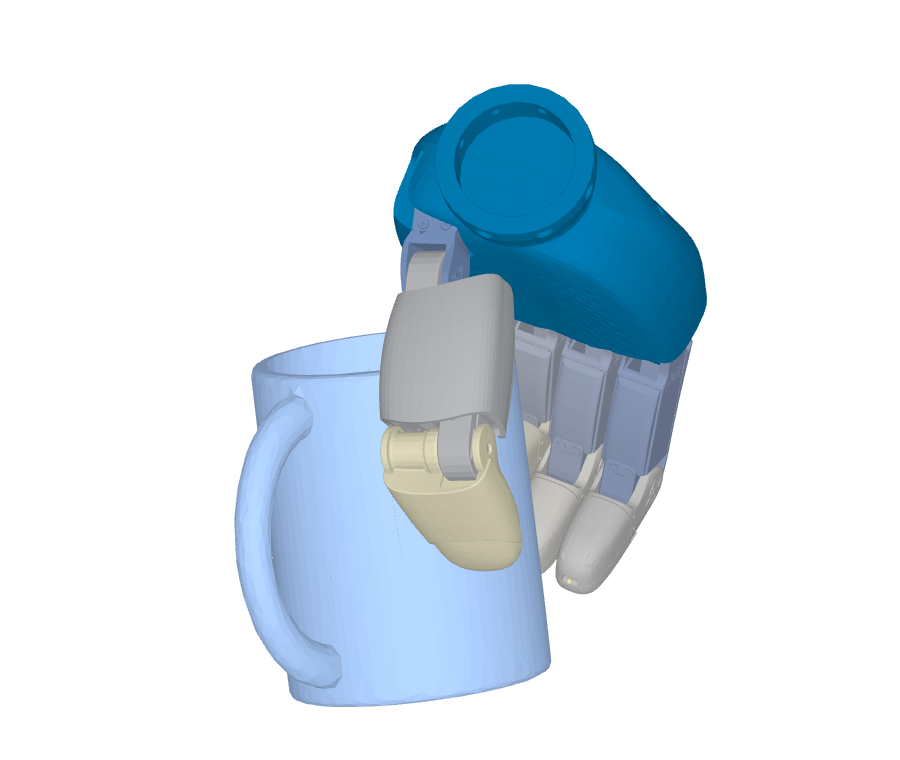} &
        \includegraphics[width=0.13\textwidth]{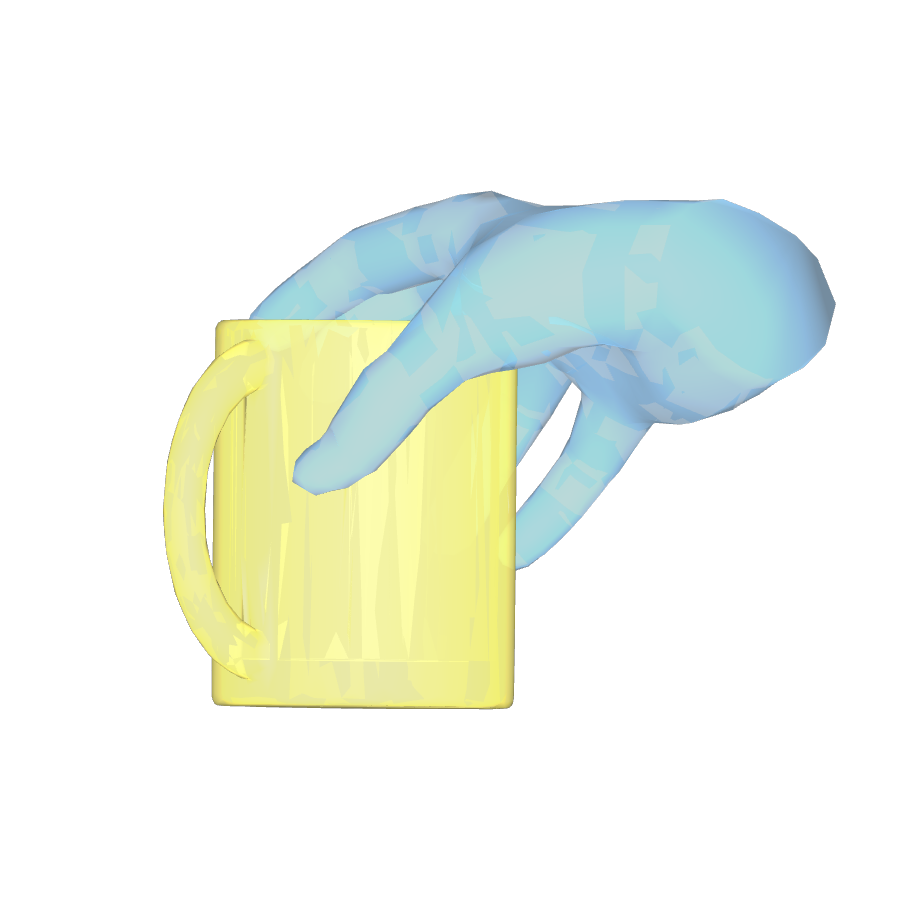} &
        \includegraphics[width=0.18\textwidth]{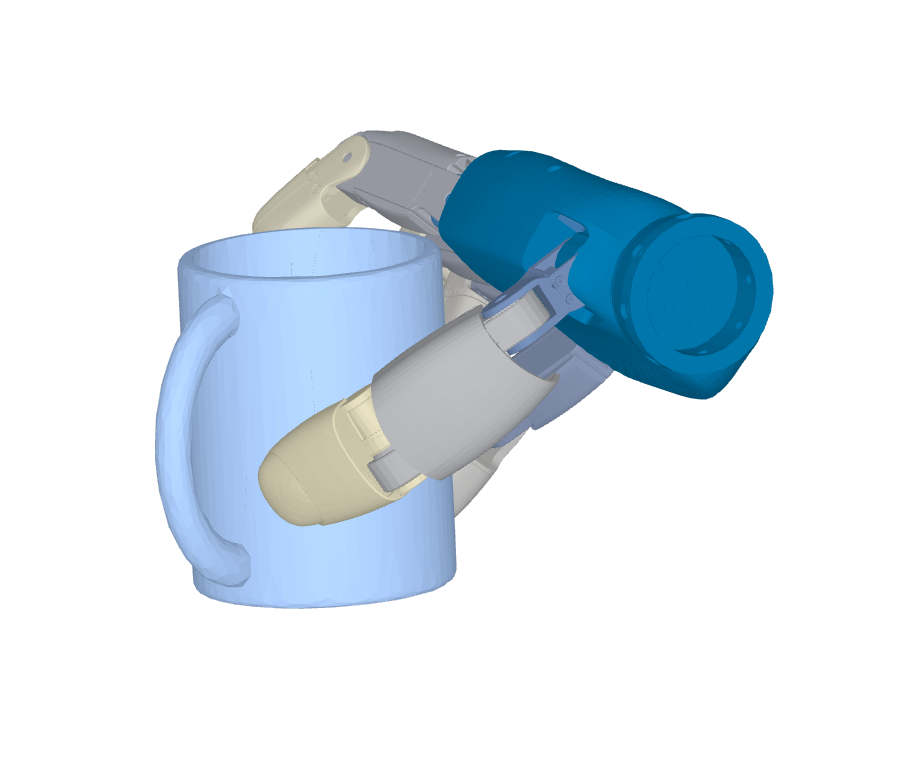} \\
        Bottle side &
        \includegraphics[width=0.13\textwidth]{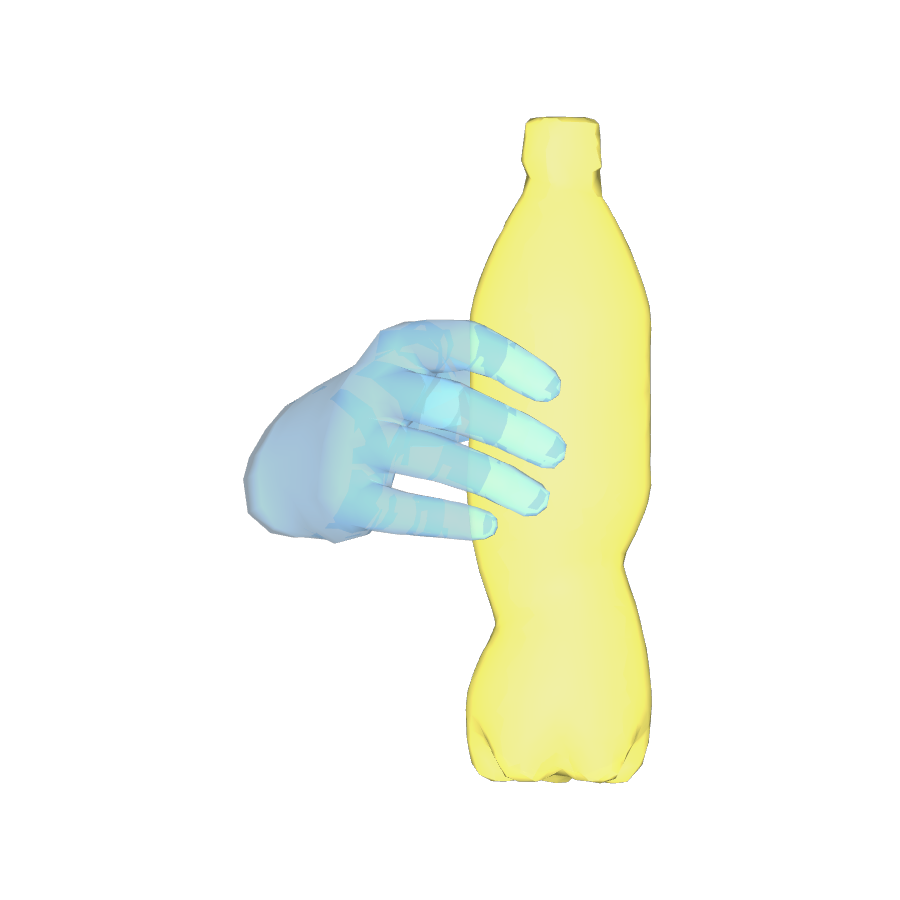} &
        \includegraphics[width=0.18\textwidth]{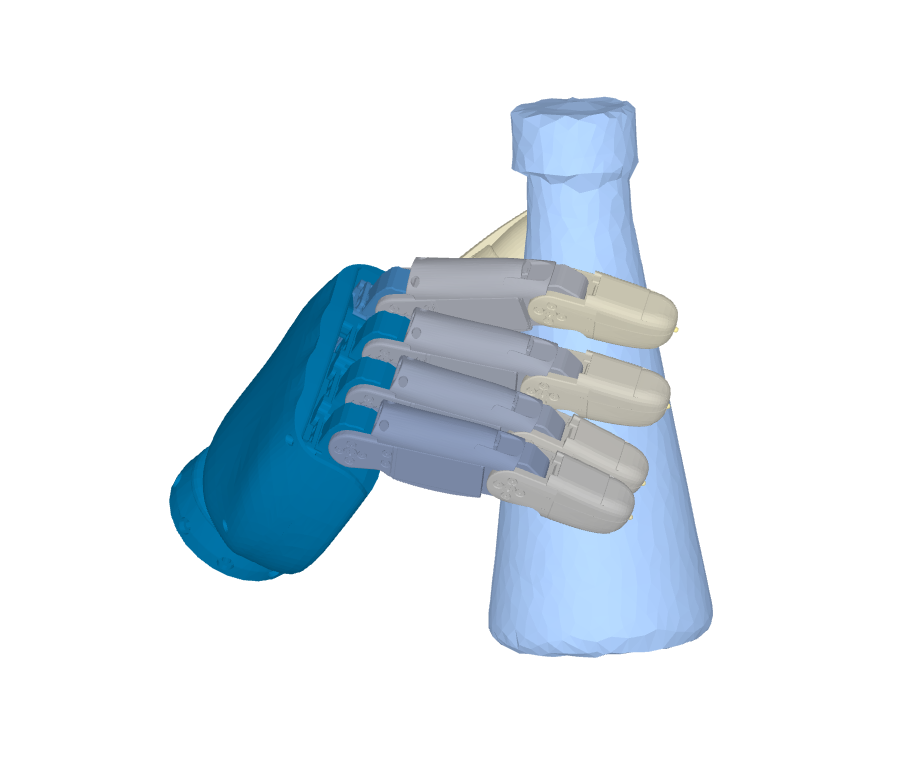} &
        \includegraphics[width=0.13\textwidth]{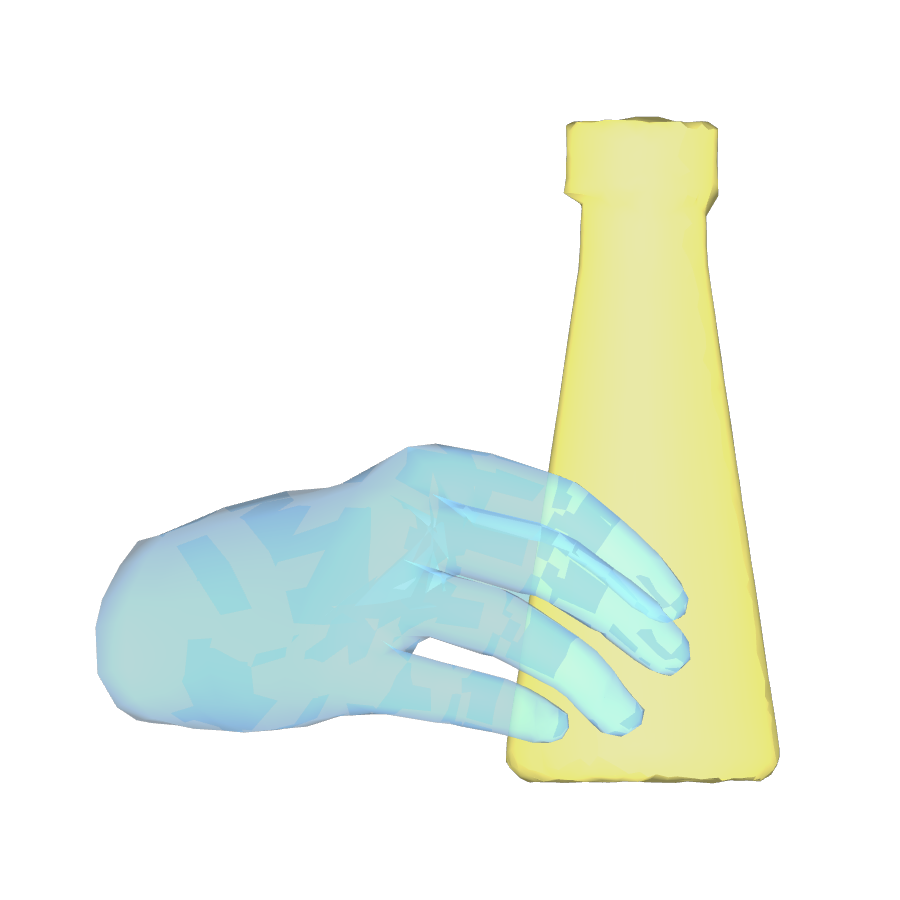} &
        \includegraphics[width=0.18\textwidth]{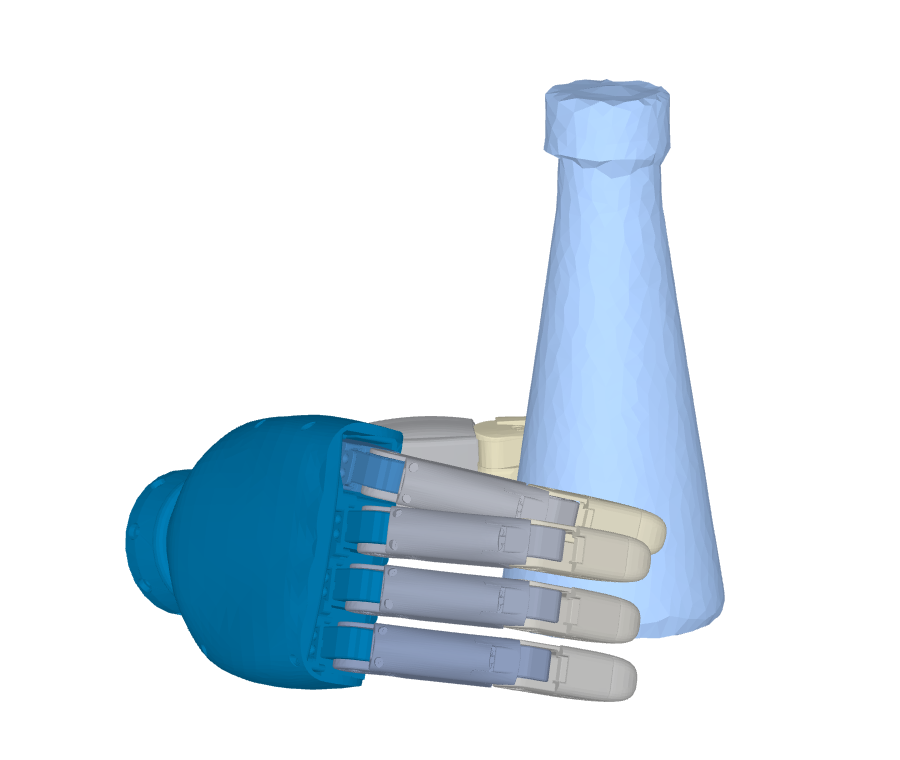} \\
        Trigger sprayer &
        \includegraphics[width=0.13\textwidth]{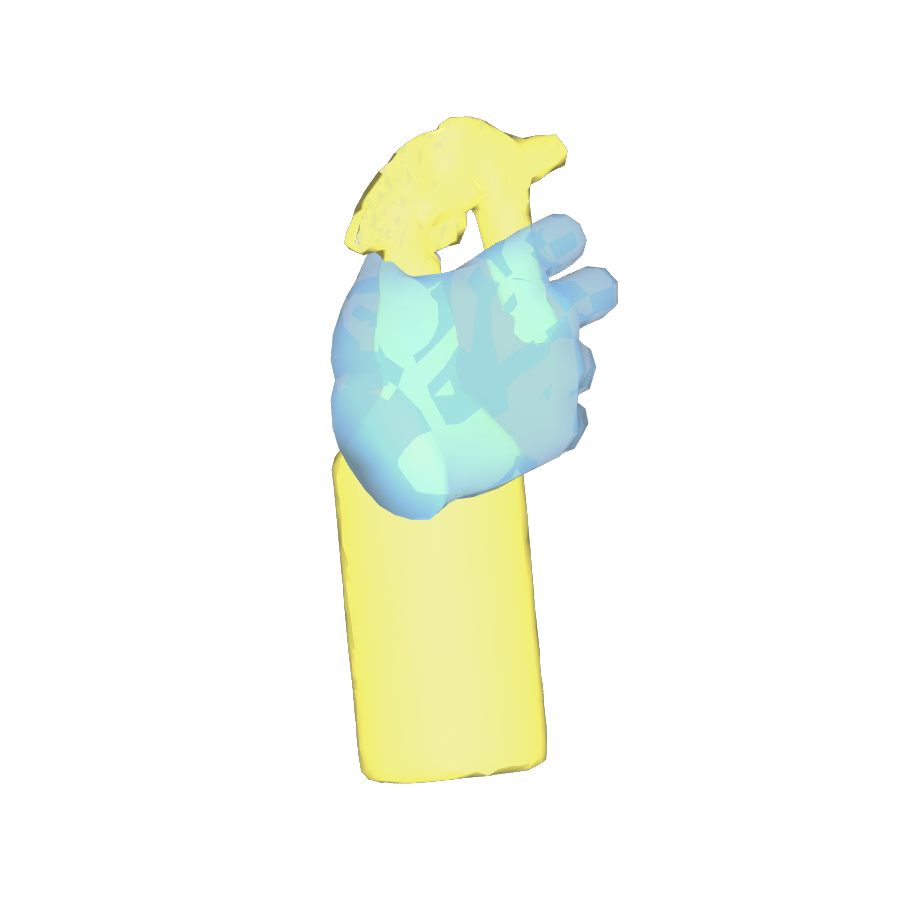} &
        \includegraphics[width=0.18\textwidth]{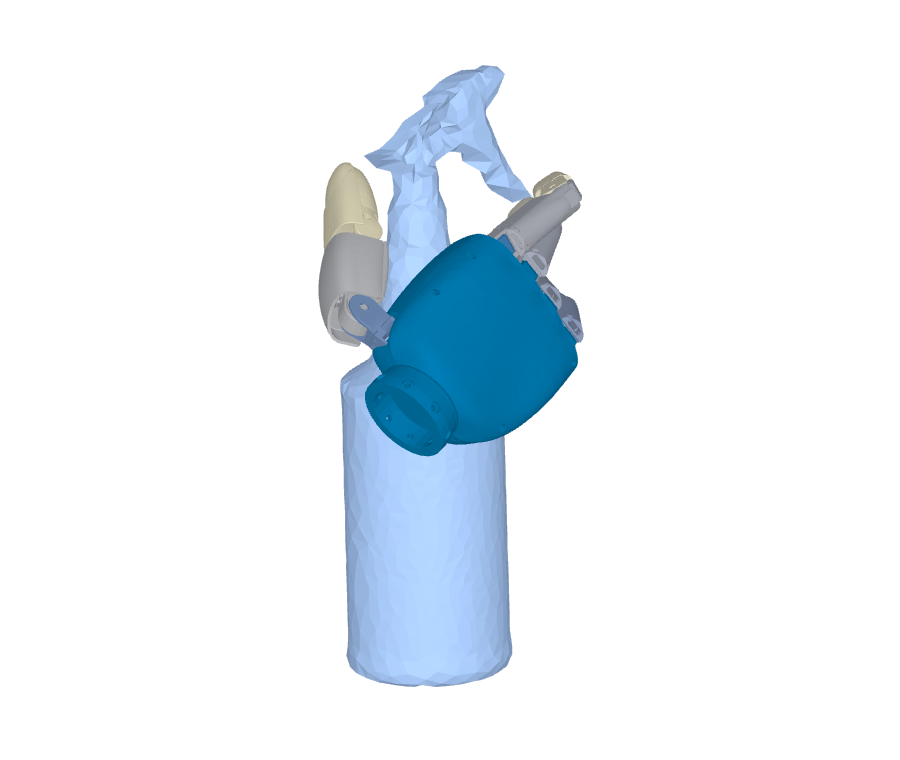} &
        \includegraphics[width=0.13\textwidth]{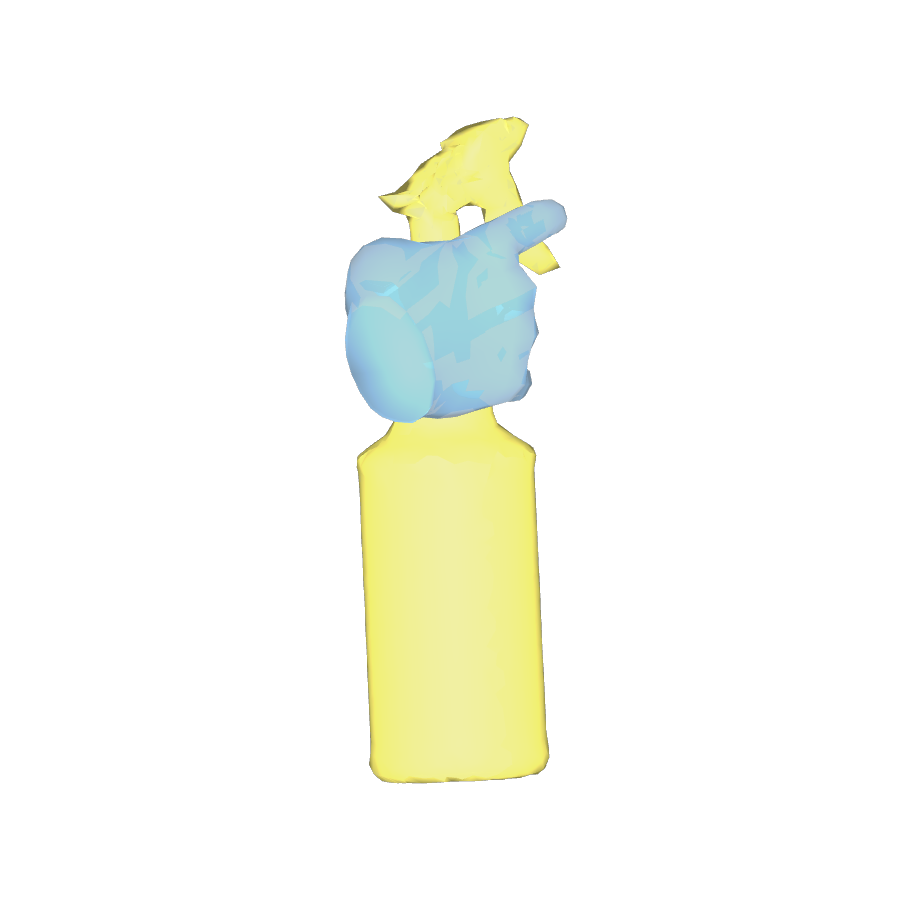} &
        \includegraphics[width=0.18\textwidth]{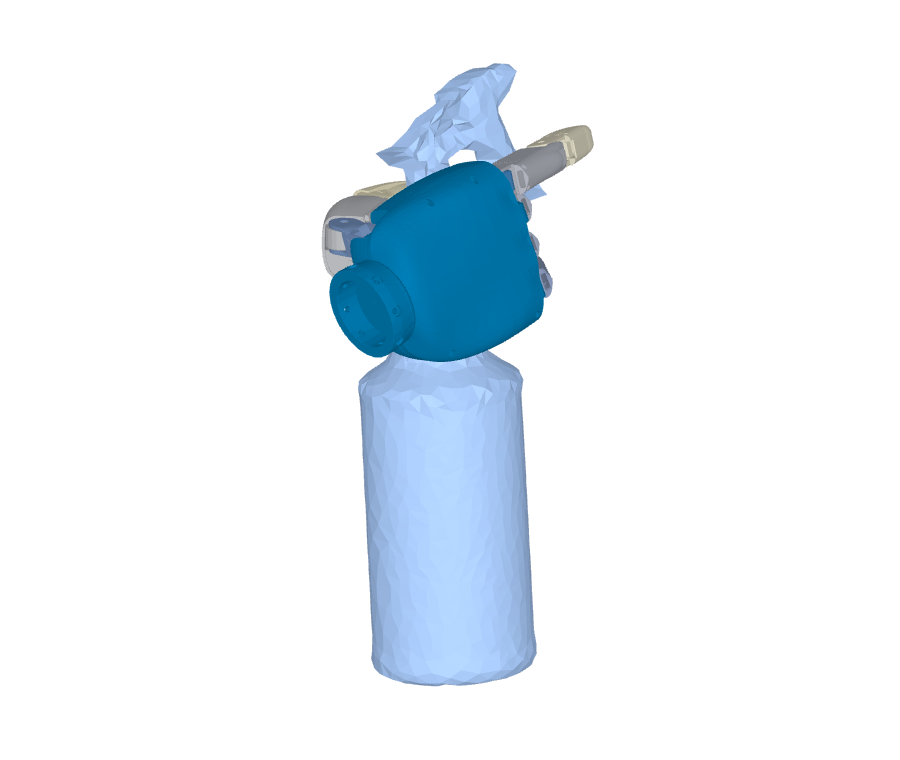} \\
        Apple$\rightarrow$camera &
        \includegraphics[width=0.13\textwidth]{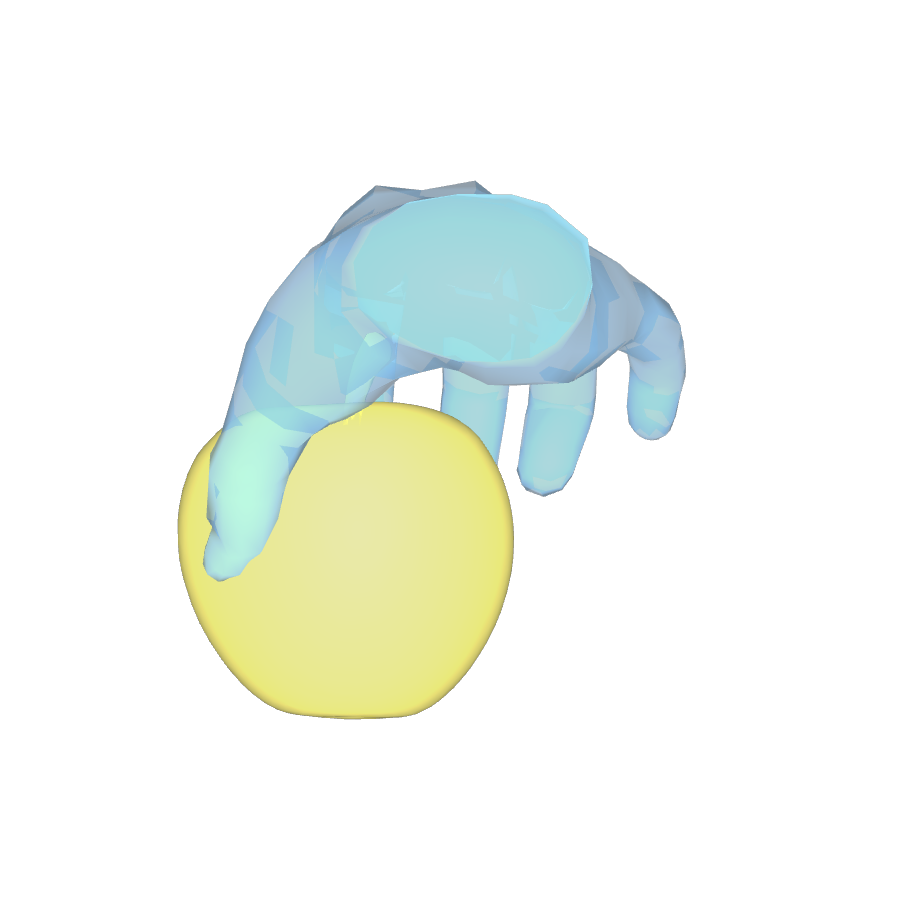} &
        \includegraphics[width=0.18\textwidth]{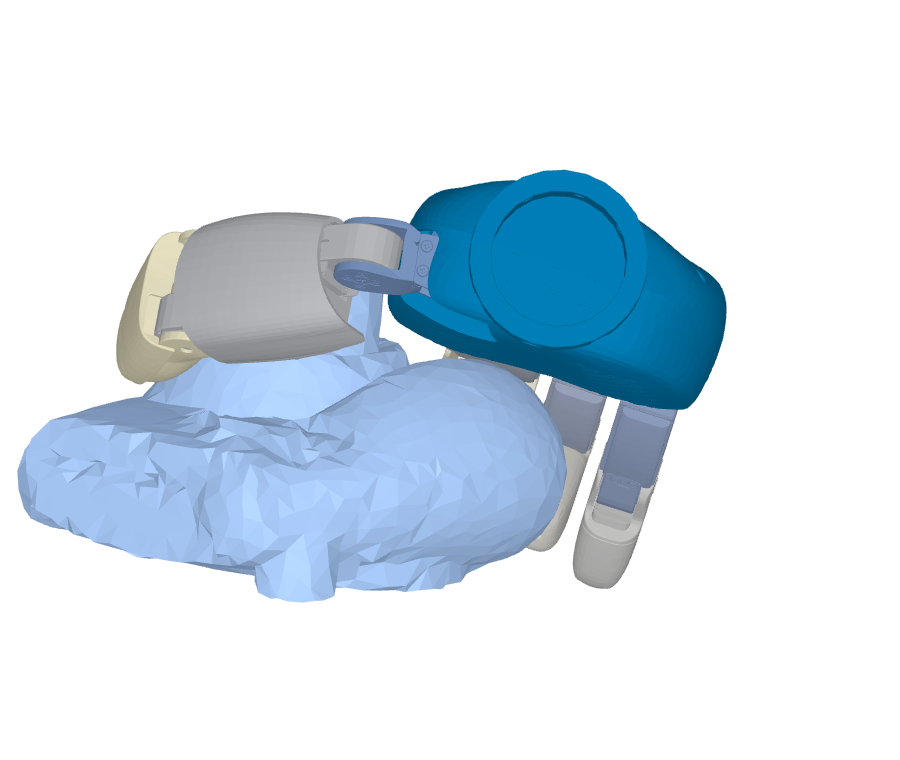} &
        \includegraphics[width=0.13\textwidth]{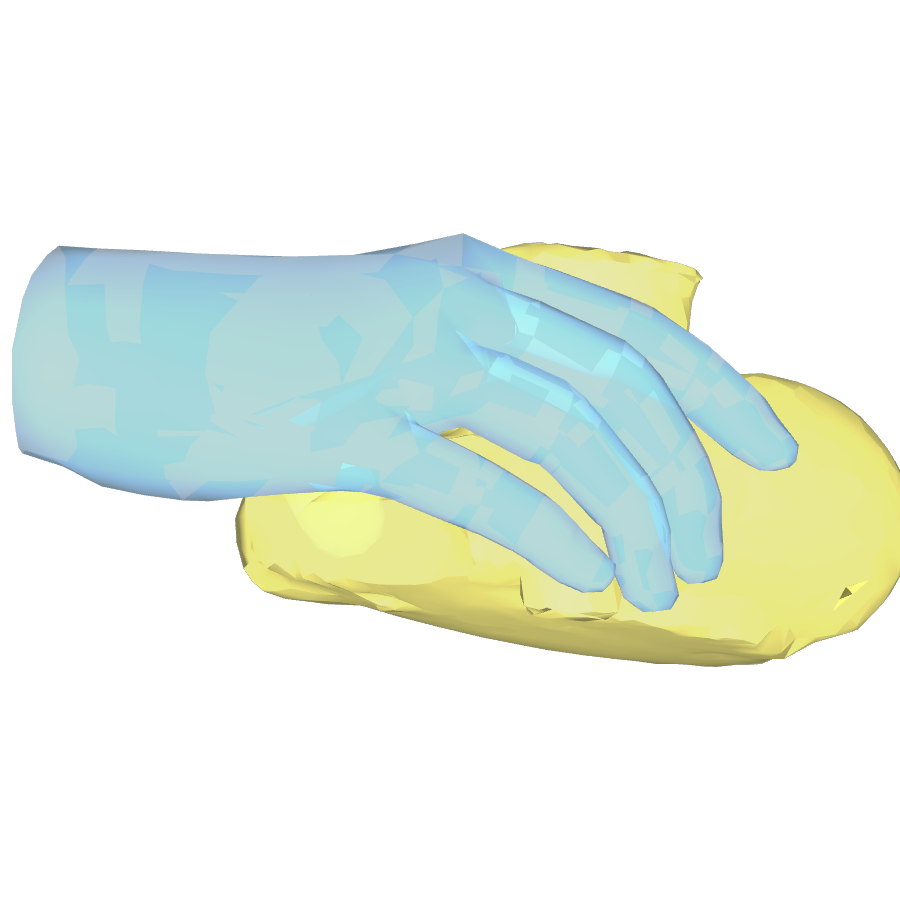} &
        \includegraphics[width=0.18\textwidth]{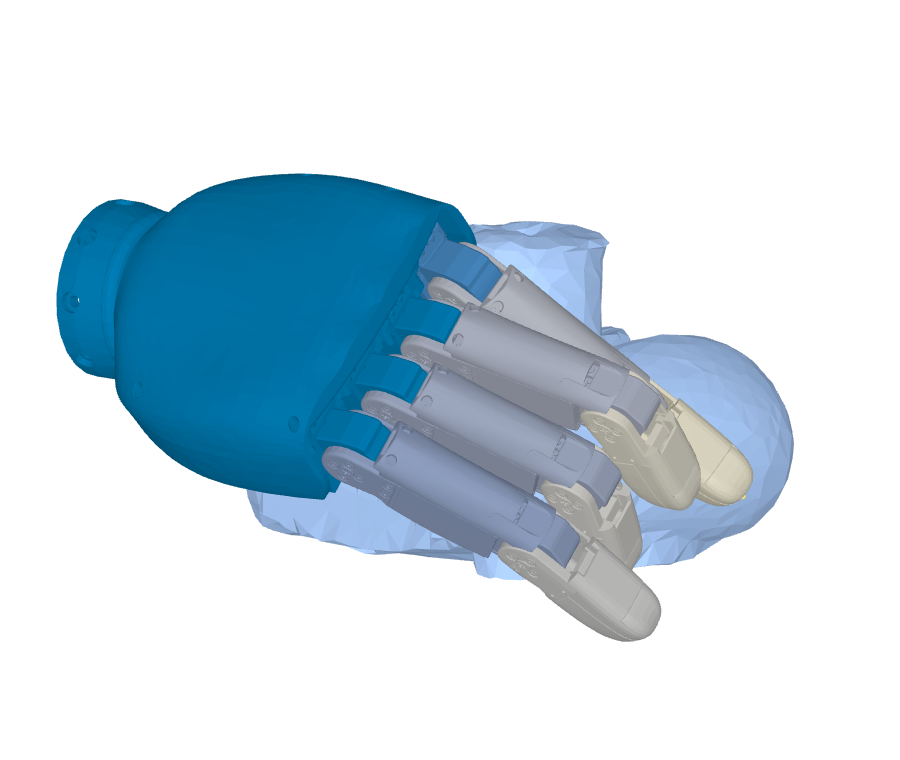} \\
        Hammer$\rightarrow$wineglass &
        \includegraphics[width=0.13\textwidth]{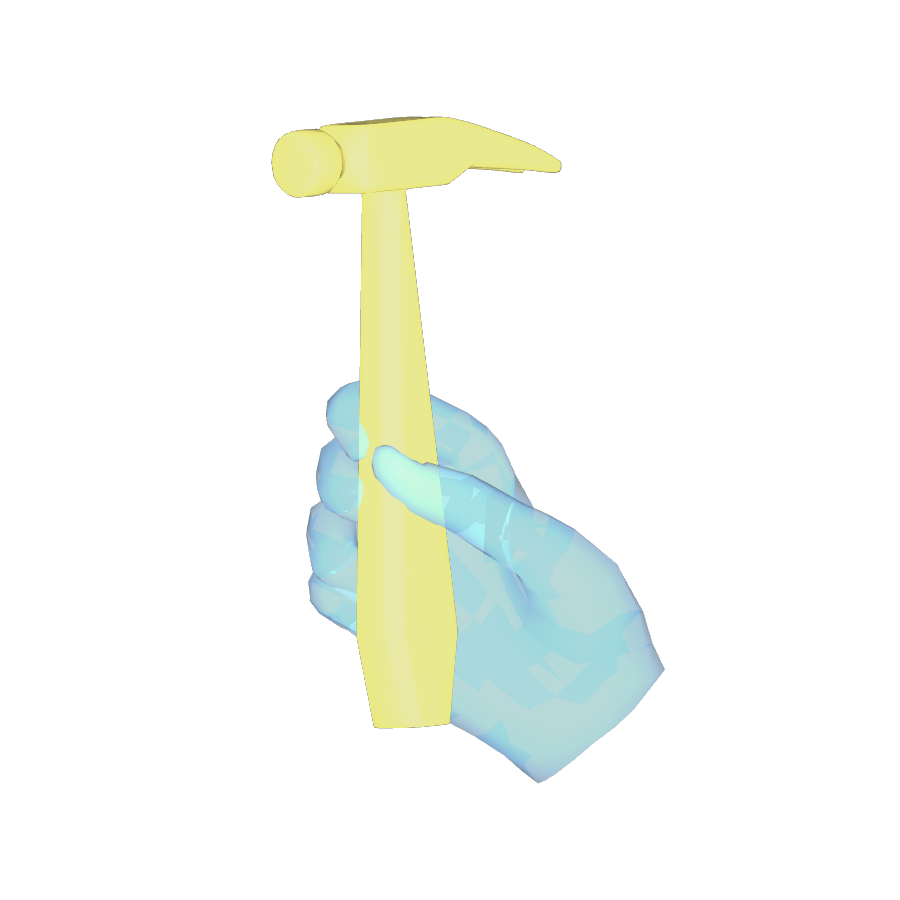} &
        \includegraphics[width=0.18\textwidth]{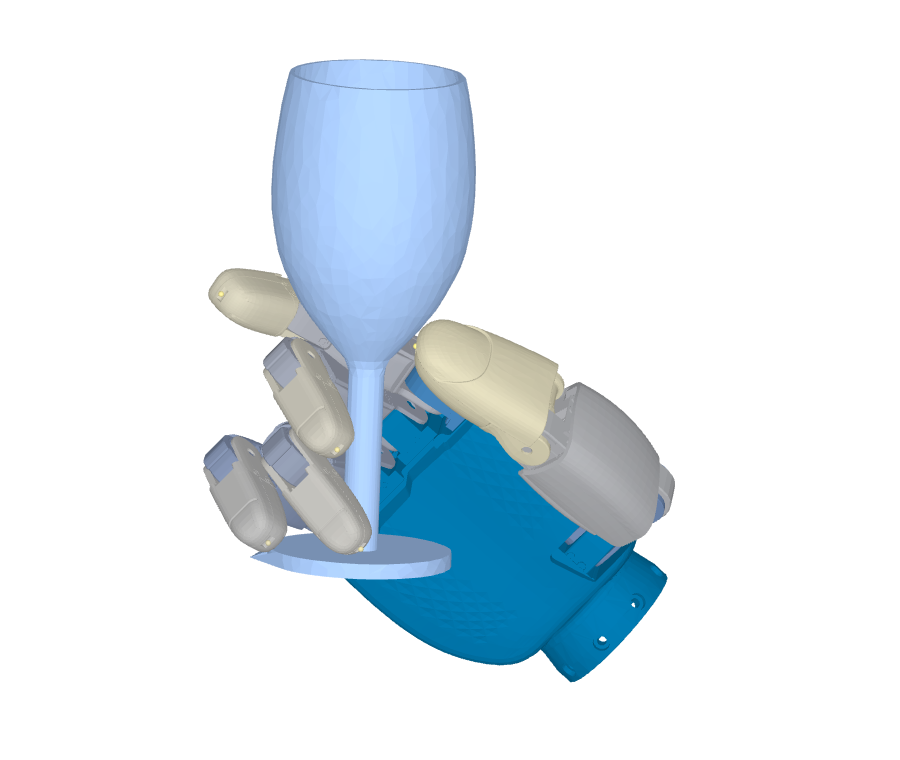} &
        \includegraphics[width=0.13\textwidth]{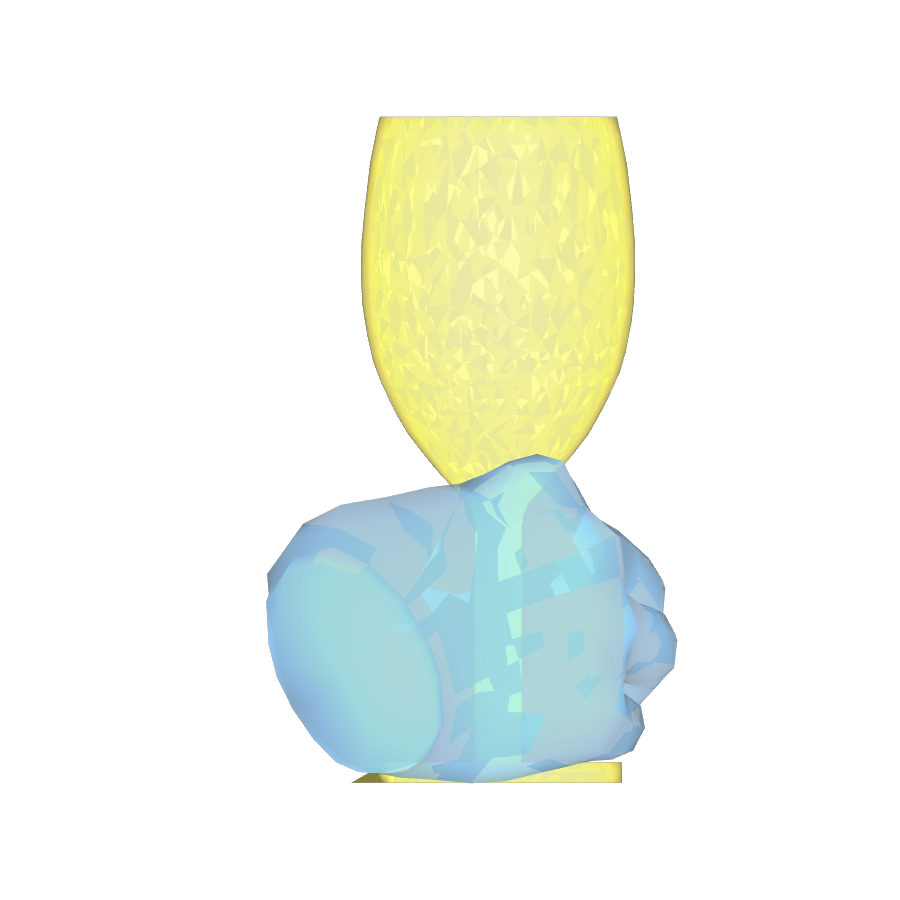} &
        \includegraphics[width=0.18\textwidth]{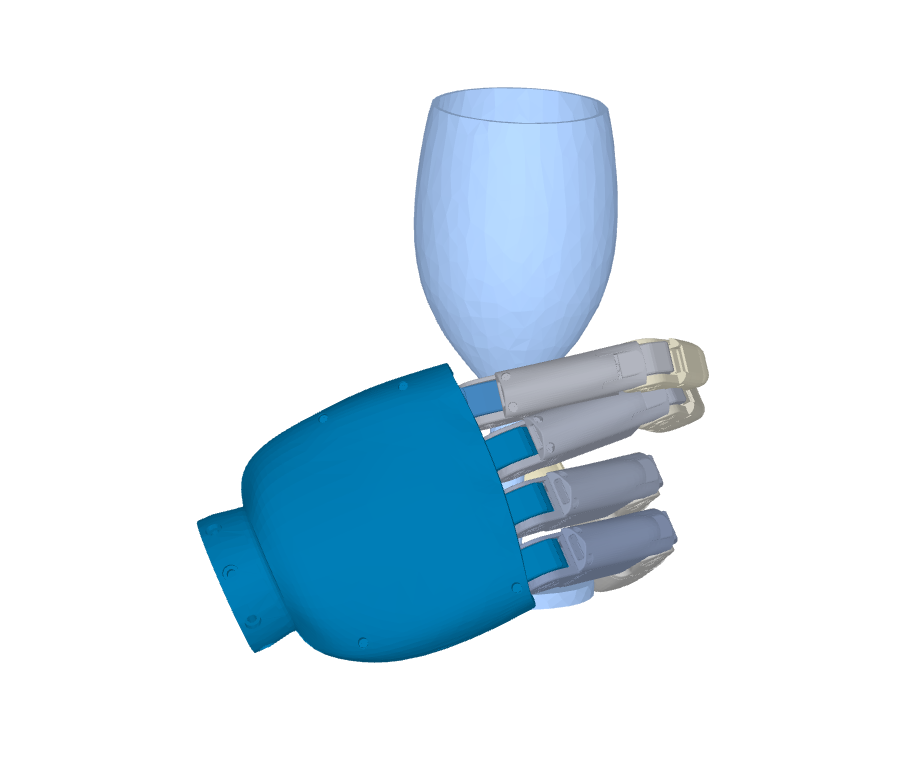} \\
        Apple$\rightarrow$mouse &
        \includegraphics[width=0.13\textwidth]{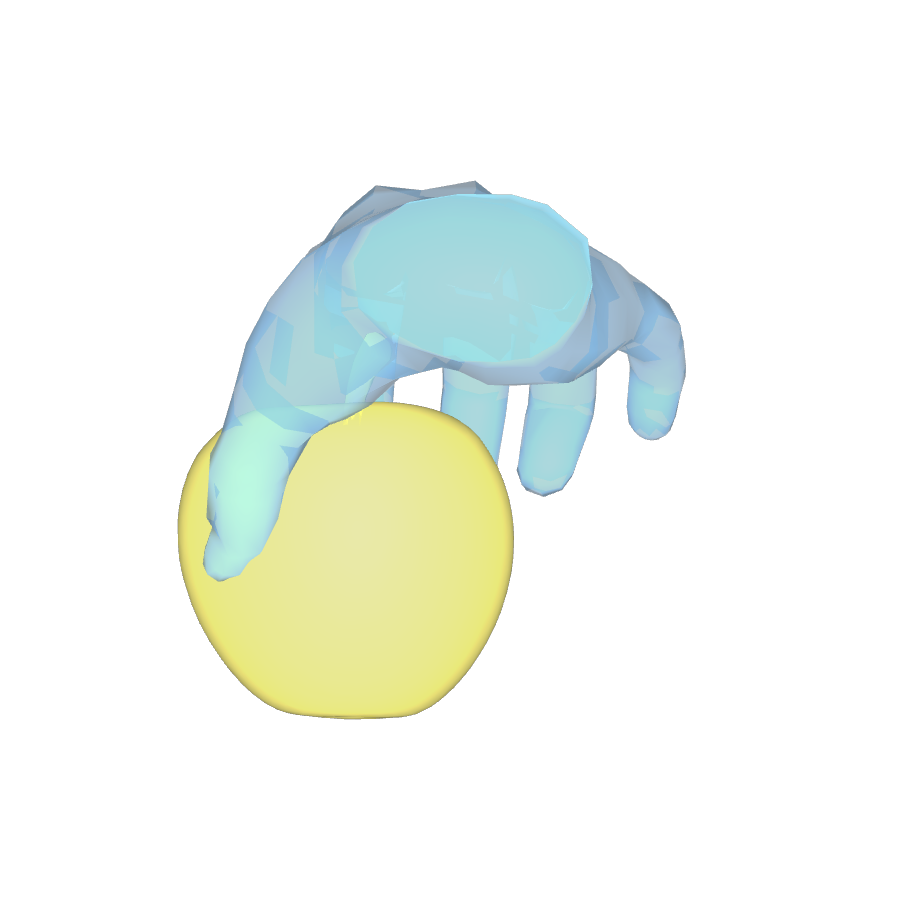} &
        \includegraphics[width=0.18\textwidth]{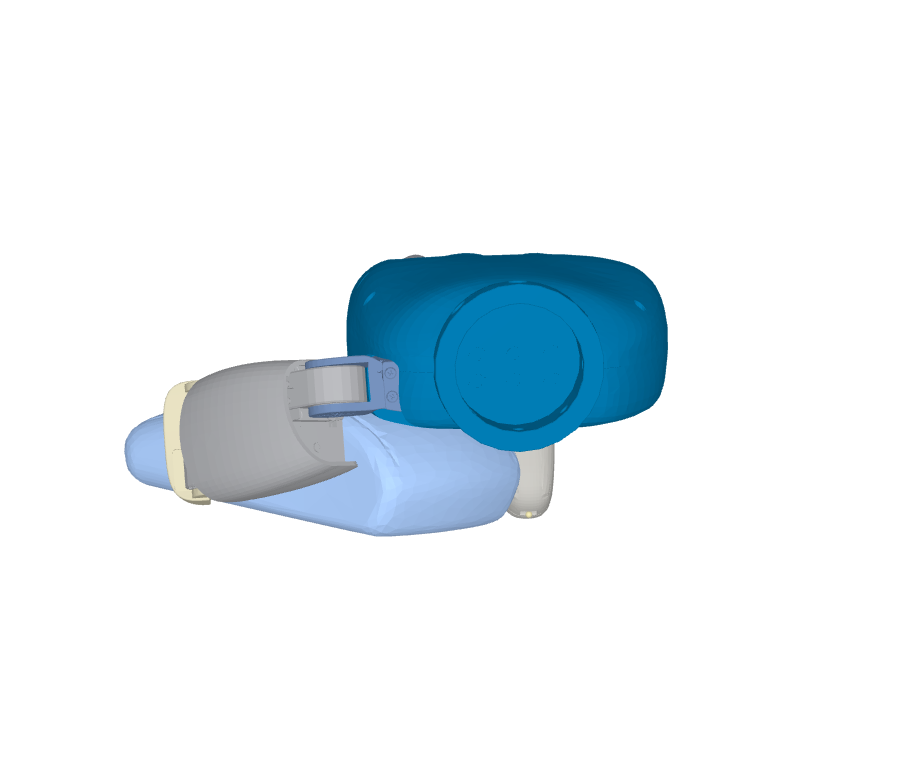} &
        \includegraphics[width=0.13\textwidth]{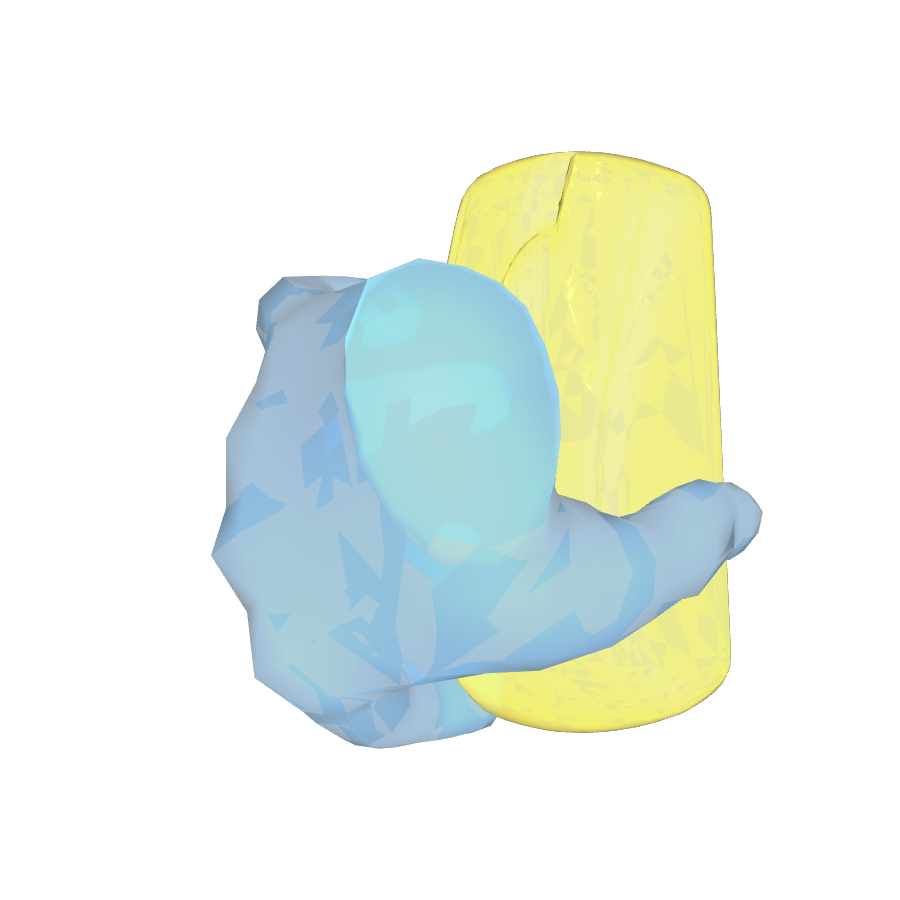} &
        \includegraphics[width=0.18\textwidth]{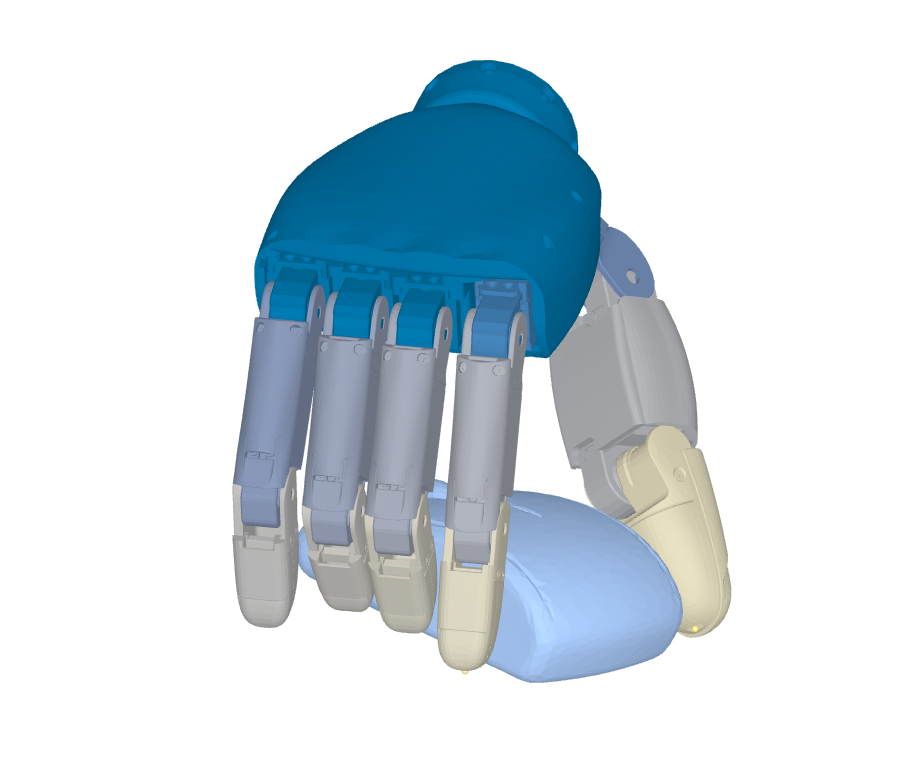} \\
    \end{tabular}
    \caption{Transferred and exact-target interaction inputs used in the
    template-transfer comparison. Each row holds the target object and downstream
    RL task fixed; only the source interaction template changes.}
    \label{fig:suppv2-transfer-template-grid}
\end{figure*}

The scale study fixes one transferred target mesh within each object category
and changes only its isotropic scale. The human template object and target mesh
are different instances in every scale-reuse family. The transfer study instead
compares a cross-instance or cross-category template with an exact-target
template while holding the target object and reach--grasp--lift task fixed. Table \ref{tab:suppv2-reuse-tasks} covers additional details, and Figure \ref{fig:suppv2-scale-template-grid} and \ref{fig:suppv2-transfer-template-grid} show the templates used visually.

\begin{table}[ht]
    \centering
    \scriptsize
    \caption{Task construction for the reusability experiments. All references
    contain 20 reach, 5 grasp, 20 lift, and 5 final-hold frames.}
    \label{tab:suppv2-reuse-tasks}
    \begin{tabular}{@{}p{0.12\linewidth}p{0.15\linewidth}p{0.24\linewidth}cccp{0.25\linewidth}@{}}
        \toprule
        \textbf{Study} & \textbf{Conditions} & \textbf{Interaction families} & $H$ & $K$ & $T$ & \textbf{Success criterion} \\
        \midrule
        Scale reuse & $0.70\times$, $0.85\times$, $1.00\times$, $1.25\times$ & Bottle side/top; mug handle/side/top. One fixed ShapeNet target per category is rescaled. & 1 & 2 & 50 & Final vertical displacement is within 3~cm of the reference lift; early termination is failure. \\
        Template transfer & Transferred versus exact-target template & Mug top, bottle side, trigger spraying, apple$\rightarrow$camera, hammer$\rightarrow$wineglass, apple$\rightarrow$mouse & 1 & 2 & 50 & Final vertical displacement is within 3~cm of the reference lift; early termination is failure. \\
        \bottomrule
    \end{tabular}
\end{table}

\clearpage
\subsection{Controllability}
\label{sec:suppv2-task-control}

The controllability experiments intervene on one aspect of the task
specification at a time. Table~\ref{tab:suppv2-control-tasks} records both the
direct template budget and the number of derived keyframes used to compile the
reference.

\begingroup
\scriptsize
\setlength{\tabcolsep}{2pt}
\begin{longtable}{@{}p{0.10\linewidth}p{0.15\linewidth}p{0.14\linewidth}rrrrp{0.30\linewidth}@{}}
    \caption{Task and reference details for the controllability experiments.}
    \label{tab:suppv2-control-tasks}\\
    \toprule
    \textbf{Condition} & \textbf{Human source} & \textbf{Robot keyframes} & $H$ & $K$ & $T$ & $H/T$ & \textbf{Success criterion} \\
    \midrule
    \endfirsthead
    \toprule
    \textbf{Condition} & \textbf{Human source} & \textbf{Robot keyframes} & $H$ & $K$ & $T$ & $H/T$ & \textbf{Success criterion} \\
    \midrule
    \endhead
    Cup pose: $45^\circ$, $90^\circ$, $120^\circ$ & OakInk cup top grasp & Retargeted grasp; derived lift and terminal tilt & 1 & 3 & 59 & 1.7\% & Final cup position and orientation are within 3~cm and $15^\circ$ of the requested pose. \\
    Cube: pick-and-place & OakInk apple top grasp reused on a block & Retargeted grasp; derived lift, hover, place, release & 1 & 5 & 54 & 1.9\% & Final pose is within 3~cm and $30^\circ$ of the goal, and maximum center-height increase exceeds 2~cm. \\
    Cube: push & Same source, block, and terminal goal as pick-and-place & Retargeted contact; derived push end and release & 1 & 3 & 54 & 1.9\% & Final pose is within 3~cm and $30^\circ$ of the goal, and maximum center-height increase is at most 2~cm. \\
    Hammer: fast & OakInk hammer-handle grasp & Retargeted grasp; lift, ready, three arc points, strike & 1 & 7 & 78 & 1.3\% & Hammer--peg contact occurs within the authored strike window; early termination is failure. \\
    Hammer: slow & Same grasp and seven robot keyframes as fast & Same keyframes; swing phases expanded & 1 & 7 & 93 & 1.1\% & Same phase-windowed contact criterion as the fast condition. \\
    Bottle: side/top & OakInk bottle side and top grasps & Retargeted grasp and derived lift & 1 & 2 & 50 & 2.0\% & Final vertical displacement is within 3~cm of the reference lift. Interaction adherence is reported separately. \\
    \bottomrule
\end{longtable}
\endgroup


\subsection{Ablations}
\label{sec:suppv2-task-ablations}

The component ablations span both dense demonstration-like references and
sparse stage-based references. All policy variants for a task use the same
underlying object path, template budget, reference horizon, and evaluation
protocol.

\begingroup
\scriptsize
\setlength{\tabcolsep}{2pt}
\begin{longtable}{@{}p{0.10\linewidth}p{0.155\linewidth}p{0.14\linewidth}rrrrp{0.295\linewidth}@{}}
    \caption{Task and reference details for the four-task ablation suite.}
    \label{tab:suppv2-ablation-tasks}\\
    \toprule
    \textbf{Task} & \textbf{Human source} & \textbf{Robot keyframes} & $H$ & $K$ & $T$ & $H/T$ & \textbf{Success criterion} \\
    \midrule
    \endfirsthead
    \toprule
    \textbf{Task} & \textbf{Human source} & \textbf{Robot keyframes} & $H$ & $K$ & $T$ & $H/T$ & \textbf{Success criterion} \\
    \midrule
    \endhead
    Open notebook & ARCTIC notebook-opening sequence & 37 consecutive single-hand retargeted frames & 37 & 37 & 58 & 63.8\% & Final notebook articulation is within $15^\circ$ of the reference target. \\
    Stack RGB blocks & One OakInk apple top grasp reused for three blocks & Per block: grasp, lift, hover, place, release & 3 instances & 15 & 148 & 2.0\% & Complete, aligned, released stack with intended support and 1~s passive stability. \\
    Banana handover & ARCTIC bimanual handover sequence transferred from a ketchup bottle to a banana & 30 paired bimanual retargeted frames & 30 paired & 30 paired & 37 & 81.1\% & Banana center is at least 5~cm above the table (not dropped), and left hand is supporting the banana's weight. \\
    Mug grasp change & OakInk top-diagonal and side mug grasps & Two retargeted grasps; derived lift and translation & 2 & 4 & 75 & 2.7\% & Functional lift: final rise is $0.12\pm0.03$~m. Grasp validity additionally requires an ordered top-to-side grasp similarity. \\
    \bottomrule
\end{longtable}
\endgroup

Stacking success rates and DTW in Table~\ref{tab:ablations} use separate
50-episode-per-seed evaluation batches of the same checkpoints. The batch used
for DTW achieved 82.0\% success for the full method and 30.0\% for initial-only
resets. DTW uses strict-success trajectories from that batch and excludes the
passive stability window.


\subsection{Real-World Rollouts}
\begin{figure}
    \centering
    \includegraphics[width=\columnwidth]{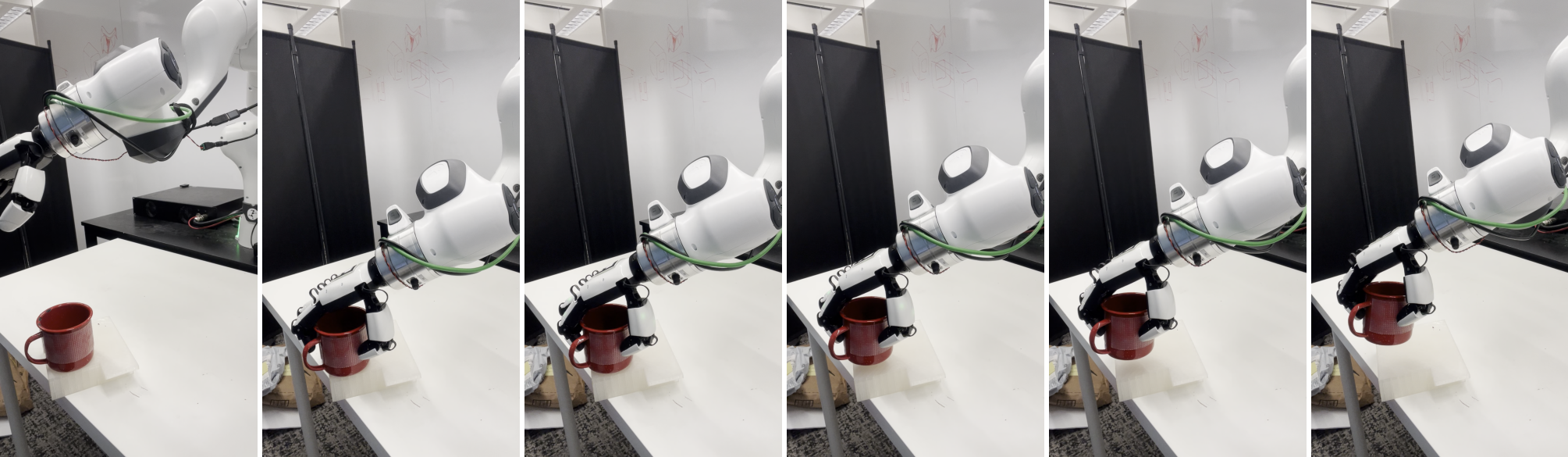}
    \vspace{2pt}

    \includegraphics[width=\columnwidth]{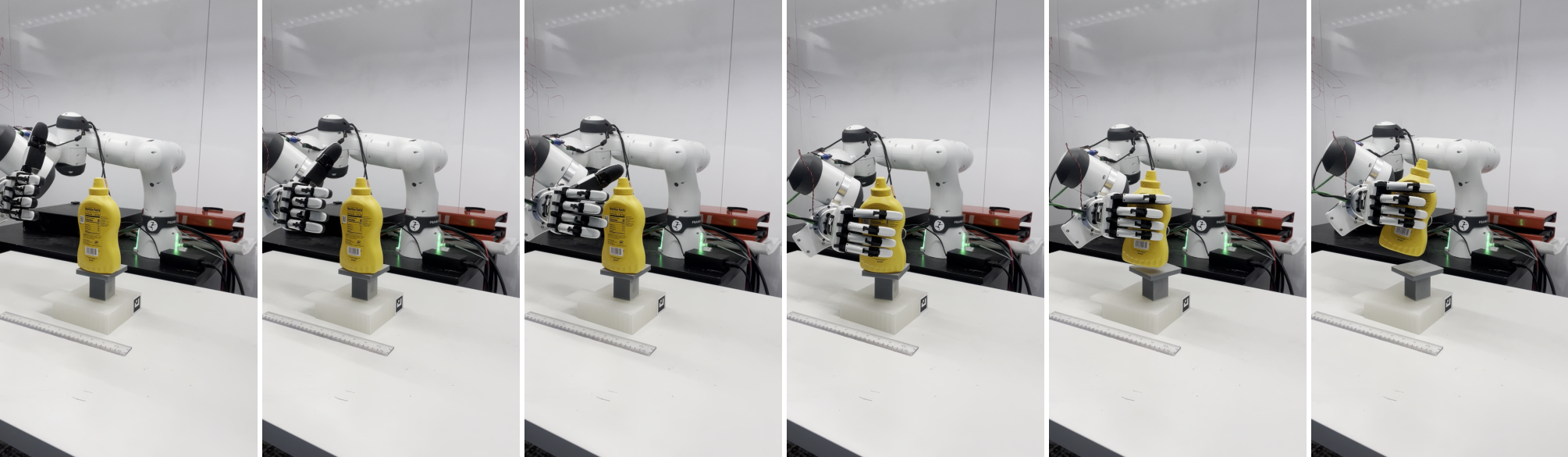}
    \vspace{2pt}

    \includegraphics[width=\columnwidth]{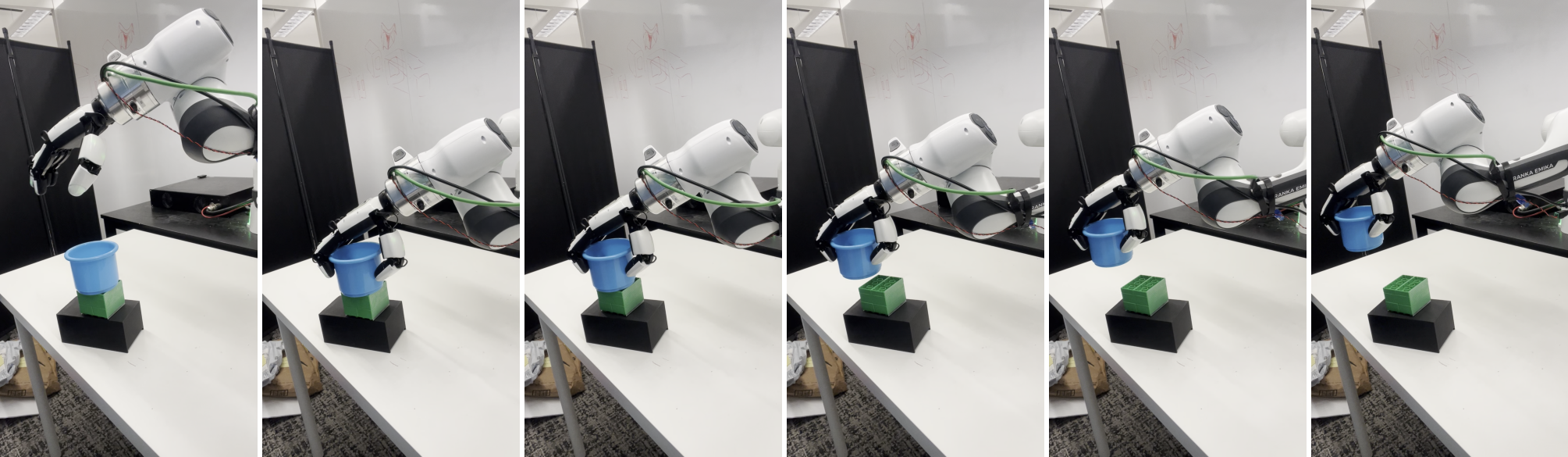}
    \vspace{2pt}

    \includegraphics[width=\columnwidth]{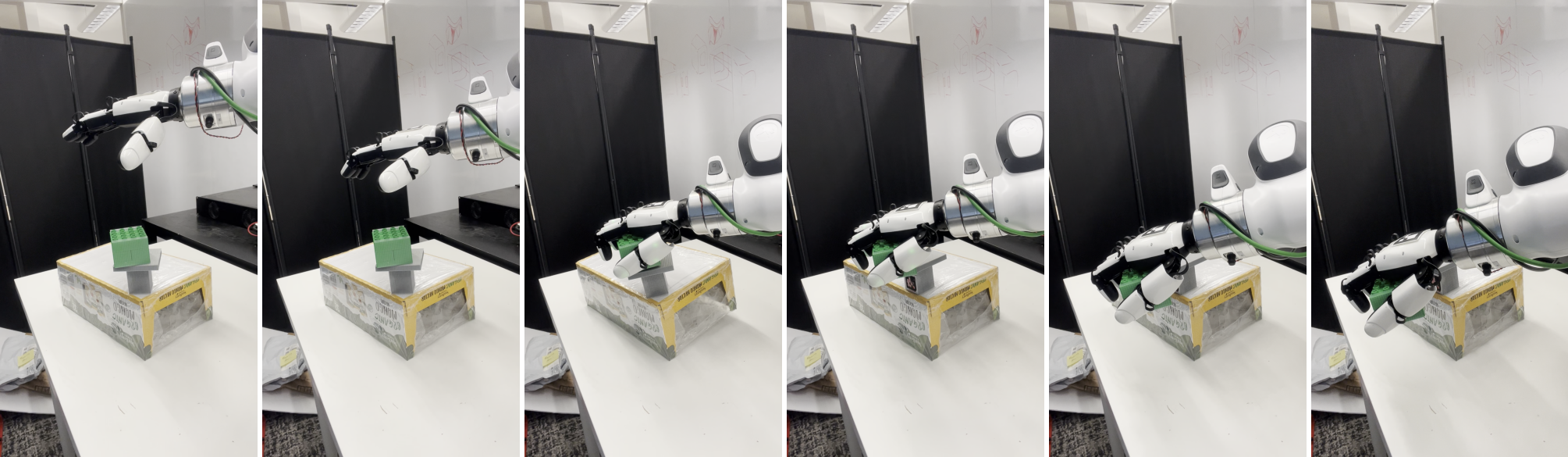}
    \caption{Real-world grasp-and-lift rollouts for the mug, mustard bottle,
    cup, and block. Each row shows uniformly sampled frames from approach to
    lift.}
    \label{fig:real_world_grasp_sequences}
\end{figure}

Table \ref{tab:supp-real-settings} and Figure \ref{fig:real_world_grasp_sequences} provide details on the real-world setup and example rollouts, respectively.

\begin{table}
    \centering
    \small
    \caption{Real-world deployment and evaluation settings.}
    \label{tab:supp-real-settings}
    \begin{tabular}{@{}p{0.28\linewidth}p{0.66\linewidth}@{}}
        \toprule
        \textbf{Setting} & \textbf{Value} \\
        \midrule
        Robot & Franka arm with XHand \\
        Sensor & Intel RealSense RGB-D camera \\
        Policy input & Filtered table, object, and robot-hand point clouds \\
        Policy output & delta-wrist pose and finger joints (same as simulation). delta-wrist pose is further processed to 6-DoF wrist actions, and inverse kinematics produces Franka joint targets \\
        Success criterion & Object held in the air for $\sim1$ minute.  \\
        \bottomrule
    \end{tabular}
\end{table}